\documentclass[ejs]{imsart}

\RequirePackage[OT1]{fontenc}
\RequirePackage{amsthm,amsmath}
\RequirePackage[numbers]{natbib}
\RequirePackage[colorlinks,citecolor=blue,urlcolor=blue]{hyperref}

\pubyear{2014}
\volume{8}
\issue{2}
\firstpage{3004}
\lastpage{3030}
\arxiv{1305.0319}
\doi{10.1214/14-EJS981}

\startlocaldefs
\numberwithin{equation}{section}
\theoremstyle{plain}

\endlocaldefs

\usepackage{graphicx}
\usepackage{amsmath}
\usepackage{amssymb}
\usepackage{algorithm}
\usepackage{algorithmic}
\usepackage{multirow}
\usepackage{color}
\usepackage{hyperref}

\graphicspath{{figs1/}{pics/}{}{pics/dog_AND/}}
\DeclareGraphicsExtensions{.jpg,.pdf,.mps,.png}
\def\RR{\mathbb R}

\newcommand{\bx}{\mathbf{x}}
\newcommand{\by}{\mathbf{y}}
\newcommand{\bz}{\mathbf{z}}
\newcommand{\bA}{\mathbf{A}}
\newcommand{\bB}{\mathbf{B}}
\newcommand{\bP}{\mathbf{P}}
\newcommand{\bQ}{\mathbf{Q}}
\newcommand{\bT}{\mathbf{T}}
\newcommand{\bu}{\mathbf{u}}

\newtheorem{theorem}{Theorem}

\newtheorem{prop}{Proposition}
\newtheorem{lem}{Lemma}
\newtheorem{definition}{Definition}

\begin{document}

\begin{frontmatter}
\runtitle{Two-Round EM with Performance Guarantee}
\title{Learning Mixtures of Bernoulli Templates by Two-Round EM with Performance Guarantee
%\thanksref{T1}
}
%\thankstext{T1}{Footnote to the title with the ``thankstext'' command.}

\begin{aug}
\author{\fnms{Adrian} \snm{Barbu}, }%\thanksref{t1,t2,m1}\ead[label=e1]{abarbu@stat.fsu.edu}},
\author{\fnms{Tianfu} \snm{Wu}}%\thanksref{t1,m2}
\and
\author{\fnms{Ying Nian} \snm{Wu}}%\thanksref{t3,m1,m2}\ead[label=e2]{ywu@stat.ucla.edu}}
%\ead[label=e3]{sczhu@stat.ucla.edu}
%\ead[label=u1,url]{http://www.foo.com}}

%\thankstext{t1}{Some comment}
%\thankstext{t2}{First supporter of the project}
%\thankstext{t3}{Second supporter of the project}
\runauthor{Barbu et al.}

\affiliation{Florida State University %\thanksmark{m1} 
and UCLA}%\thanksmark{m2}}

%\address{Address of the First and Second authors\\
%Usually a few lines long\\
%\printead{e1}\\
%\phantom{E-mail:\ }\printead*{e2}}

%\address{Address of the Third author\\
%Usually a few lines long\\
%Usually a few lines long\\
%\printead{e3}\\
%\printead{u1}}
\end{aug}

\begin{abstract}
Dasgupta and Shulman \cite{dasgupta2000two} showed that a two-round variant of the EM algorithm can learn mixture of Gaussian distributions with near optimal precision with high probability if the Gaussian distributions are well separated and if the dimension is sufficiently high. In this paper, we generalize their theory to learning mixture of high-dimensional Bernoulli templates. Each template is a binary vector, and a template generates examples by randomly switching its binary components independently with a certain probability. In computer vision applications, a binary vector is a feature map of an image, where each binary component indicates whether a local feature or structure is present or absent within a certain cell of the image domain. A Bernoulli template can be considered as a statistical model for images of objects (or parts of objects) from the same category. We show that the two-round EM algorithm can learn mixture of Bernoulli templates with near optimal precision with high probability, if the Bernoulli templates are sufficiently different and if the number of features is sufficiently high. We illustrate the theoretical results by synthetic and real examples. 
\end{abstract}

%\begin{keyword}[class=MSC]
%\kwd[Primary ]{60K35}
%\kwd{60K35}
%\kwd[; secondary ]{60K35}
%\end{keyword}


\end{frontmatter}

\section{Introduction} 

During the past decades, a large number of theoretical results have been obtained for supervised learning such as classification and regression \cite{svm}. For unsupervised learning, however, relatively few theoretical results are available. A main difficulty is that the objective functions in unsupervised learning are usually non-convex and multi-modal, so the optimization algorithms usually cannot find the global optima. As a result, it is generally difficult to obtain theoretical guarantees for the performances of the unsupervised learning algorithms. A simple and typical example of unsupervised learning is clustering or learning mixture models, and a typical algorithm for fitting the mixture models is the EM algorithm \cite{dempster1977maximum}, which is a statistical counterpart of the k-means algorithm for clustering. Although the EM algorithm is simple and interpretable, and is known to converge monotonically to a local mode of the observed-data log-likelihood, little is known about its theoretical performance in terms of correctly recovering the mixture components. As such, the EM algorithm is often considered a heuristic algorithm.

A major recent advance in the theoretical understanding of the EM algorithm for fitting mixture models was made by Dasgupta and Shulman %\citet{dasgupta2000two}
\cite{dasgupta2000two}. They proposed a two-round variant of the EM algorithm that consists of only two iterations of EM: the first iteration is initialized from a number of randomly selected training examples as the centers of the Gaussian distributions, and the second iteration is carried out after pruning the clusters learned from the first iteration. They showed that the two-round EM can learn the mixture of Gaussian distributions with near optimal precision with high probability if the Gaussian distributions are well separated and if the dimensionality of the Gaussian distributions is sufficiently high. Here near optimal precision means that one can estimate the parameters of the Gaussian distributions as if the memberships of the observations are known. 

\begin{figure}[ht]
\begin{center}
%\begin{tabular}{ccc}
\includegraphics[height=.25\textwidth]{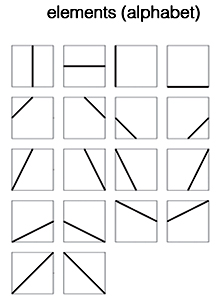} 
 \includegraphics[height=.22\textwidth]{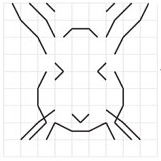} 
\includegraphics[height=.25\textwidth]{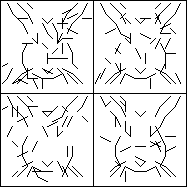} 
%\end{tabular}
\end{center}
%\vskip -3mm
\caption{Left: An alphabet of 18 sketch patterns. These sketch patterns are edge segments that connect the corners and mid-points of the sides of a squared cell. Middle: The image domain is partitioned into squared cells. Within each cell, any of the sketch patterns can be present or absent. The whole feature map can be represented by a binary vector, where each component is a binary decision on whether a certain sketch pattern in the alphabet is present or absent within a certain cell. Right: Some examples generated by the template in the middle by randomly switching the binary components with a certain probability. }
\label{fig:1}
%\vspace{-5mm}
\end{figure}

In this paper, we generalize the theory of Dasgupta and Shulman \cite{dasgupta2000two} to learning mixture of Bernoulli templates. Each template is a binary vector, and it generates examples by independently switching its binary components with a certain probability. So the observed examples are also binary vectors. This setup is a version of the latent class model of  \cite{goodman1974exploratory} restricted to binary data.
In potential applications in computer vision, a binary vector is a feature map of an image, where each binary component indicates whether a local feature or structure is present or absent within a certain cell of the image domain. Fig. \ref{fig:1} illustrates the basic idea by a synthetic example. The image domain is equally partitioned into squared cells (in the example in Fig. \ref{fig:1}, there are a total of $9 \times 9 = 81$ cells in the image domain). There is an alphabet of sketch patterns that can appear in these cells (Fig. \ref{fig:1} shows an alphabet of 18 types of sketch patterns). Each cell may contain one or more sketch patterns, so the binary vector for each image consists of $9 \times 9 \times 18$ binary components, each component indicates whether a certain sketch pattern is present or not within a certain cell. Specifically, each component is a binary decision that can be made based on local edge detection, Gabor filter responses \cite{daugman}, beamlet transformation \cite{beamlet} or a pre-trained classifier. A Gabor filter is a 2D linear filter that has a prefered orientation. Along that orientation the Gabor filter resembles a Gaussian and along the perpendicular direction it resembles the derivative of a Gaussian. It was shown in  \cite{daugman} that the Gabor filters are a good approximation of the receptive field profiles of orientation-sensitive neurons in a cat's visual cortex.

The formulation is very general. One can design any alphabet of local features or patterns, and one can use any binary detector or classifier to decide the presence or absence of these features within each cell. The whole feature map is a composition of local image features and is in the form of a binary vector, usually high dimensional (on the order of $10^3-10^5$). A template itself is a binary vector that is subject to component-wise switching or Bernoulli noise to account for the variations of the feature maps of individual images. The reason we focus on binary feature maps in this article is that they are easy to design and we do not need to make strong assumptions on their distributions such as Gaussianity. 

\begin{figure*}[ht]
\centering
\includegraphics[width=12cm]{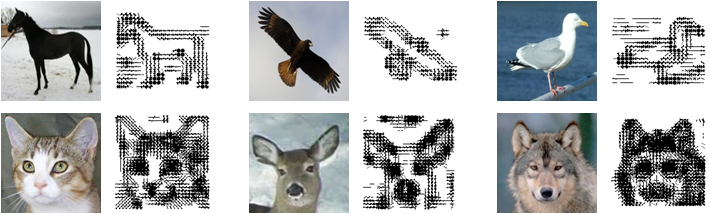}
%\vskip -3mm
\caption{Real images and their binary sketches. Each bar in the sketch image indicates the existence of a Gabor filter response above a threshold within a local cell of the image.}
\label{fig:AB}
%\vspace{-5mm}
\end{figure*}

As another illustration, Fig. \ref{fig:AB} displays some examples of real images and their binary sketches based on a simple design of image features and binary decision rules. We partition the image domain into squared cells of equal size (in these images, the cells are relatively small, ranging from $5 \times 5$ pixels to $7 \times 7$ pixels). We convolve the image with Gabor filters \cite{daugman} at 8 orientations. Within each cell, at each orientation, we pool a local maximum of the Gabor filter responses (in absolute values). If the local maximum is above a threshold, we then declare that there is a sketch within this cell at this orientation, and the sketch is depicted by a bar in the corresponding binary sketch image in Fig. \ref{fig:AB}. Clearly the sketch image captures a lot of information in the corresponding original image. 

Now back to the issue of learning mixture models by EM. We assume that there are $k$ Bernoulli templates, and each observed example is a noisy observation of one of the $k$ template. The question we want to answer is: given a number of training examples that are noisy observations of the $k$ templates, can an EM-type algorithm reliably recover these $k$ templates with high probability? The reason we are interested in this question is that it will shed light on unsupervised learning of templates of objects (or their parts) from real images, which is a crucial task for object modeling and recognition in computer vision. Many learning methods are based on fitting mixture models by EM-type algorithms, including the Active Basis model \cite{Si}. In the language of the And-Or graph \cite{Zhu_grammar} for object modeling, each template is an And-node, which is a composition of a number of sketches. The mixture of $k$ templates is an Or-node, with each template being its child node. So the mixture of the templates is an Or-And structure. The theoretical results in this paper will be useful for us to understand the learning of the Or-And structure from training images. 

To answer the above question, we shall generalize the theory of Dasgupta and Shulman \cite{dasgupta2000two} to Bernoulli distributions, and we shall show that the two-round EM algorithm can learn mixtures of Bernoulli templates with near optimal precision with high probability if the templates are sufficiently different and if the dimensions are sufficiently high. 

Generalizing the theory of \cite{dasgupta2000two} from Gaussian mixtures to the mixtures of Bernoulli distributions is far from being straightforward. The sample space is no longer Euclidean, and some results for Gaussian distributions cannot be translated directly into those for the Bernoulli models. So we have to establish a theoretical foundation that is suitable for our purpose. For example, we will need bounds on the tails of the distribution of distances between a template $\bP$ and the mean of $m$ binary vectors obtained by perturbing $\bP$ by Bernoulli noise. Similar bounds for the Gaussian case are easy to obtain because the moment generating function of $\|X\|^2$ is known when $X$ is an isotropic Gaussian.

The rest of the paper is organized as follows. Section 2 describes the two-round EM algorithm and states the main theorem. Sections 3 to 4 present theoretical results that lead to the proof of the main theorem. Section 5 illustrates the theoretical results by some experiments on synthetic and real examples. Section 6 concludes with a discussion. In the text, we shall only state the theoretical results. The proofs can be found in the appendix. 

\section{Two-round EM with performance guarantee}

\subsection{Model and algorithm} 

Let $\bP$ be a template. It is an $n$-dimensional binary vector, i.e., $\bP \in \Omega = \{0, 1\}^n$. In the example in Fig. 1, $n = 9 \times 9 \times 18=1458$. Let $\bP(s)$ be the $s$-th component of $\bP$, $s = 1, ..., n$. An example $\bx$ generated by $\bP$ is a noisy version of $\bP$, and we write $\bx \sim \bP$. Specifically, let $\bx(s)$ be the $s$-th component of $\bx$. Then $\bx(s) = \bP(s)$ with probability $1-q$, and $\bx(s) = 1 - \bP(s)$ with probability $q$, i.e., $q$ is the probability of switching a component of $\bP$, and it defines the level of Bernoulli noise. We assume that $q \in (0, 1/2)$. We also assume that the components of $\bx$ are independent given $\bP$. We call $\bP$ a Bernoulli template because it is binary and is subject to Bernoulli noise. 

Let $\{\bP_i, i = 1, ..., k\}$ be  $k$ Bernoulli templates with mixture weights $\{w_i, i = 1, ..., k\}$. We assume that $k$ is given. Otherwise, $k$ can be determined by some model selection criteria such as BIC \cite{Raftery,BIC}. Let $\bx_1, ..., \bx_m$ be $m$ noisy observations of these $k$ templates, where the noise level is $q$.  The probability that $\bx_j$ is generated by $\bP_i$ is $w_i$, and we let $w_{min}=\min_{i=1,...,k} w_i$. We define $\mu_i$ to be the expectation of the examples generated by $\bP_i$, i.e., $\mu_i = E[\bx_i]$ where $\bx_i \sim \bP_i$. Let $S_i$ be the set of examples coming from the template $\bP_i$.

For two $n$-dimensional vectors $\bP$ and $\bQ$, let $D(\bP,\bQ)=\sum_{s=1}^{n} |\bP(s)-\bQ(s)|$ be the $\ell_1$ distance between $\bP$ and $\bQ$. Let $c_{ij}$ be the separation between $\bP_i$ and $\bP_j$, i.e., $D(\bP_i, \bP_j) = d_{ij} = n c_{ij}$. 
%\vspace{-3mm}
\begin{definition} The mixture is called $c$-separated if $\min_{ij}c_{ij}= c$. 
\end{definition}
%\vspace{-3mm}

We shall show that if the separation $c$ is sufficiently large, then the two-round EM algorithm will reliably recover $\{\bP_i, i = 1, ..., k\}$. 

We use the notation $\bT_i$ to denote the estimated $\bP_i$. In the two-round EM, the first round initializes $\{\bT_i^{(0)}, i = 1, ..., l\}$ to be $l$ randomly selected training examples. The initial number of clusters, $l$, is greater than the true number $k$. Specifically, we let $l=\frac{4}{w_{min}}\ln \frac{2}{\delta w_{min}}$, where $\delta$ is the confidence parameter, i.e., with probability $1-\delta$, the algorithm will succeed in recovering the mixture components. According to the coupon collector problem, the $l$ examples cover all the $k$ clusters with high probability. We estimate the Bernoulli noise level $q_0$ so that $q_0(1-q_0) = \min_{ij} D(\bT_i^{(0)}, \bT_j^{(0)})/2n$ based on the statistics of distances between examples derived in Prop. \ref{prop:eps}. Then we run one more iteration of EM. 

\begin{algorithm}[h!]
   \caption{\bf Two-round EM for Learning Bernoulli Templates}
   \label{alg:learnem}
\begin{algorithmic}
   \STATE {\bfseries Input:} Examples $\bx_1,...,\bx_m\in \Omega$, $m\geq l$
   \STATE {\bfseries Output:} Templates $\bT_i, i=1,..,k$
   \STATE[1] Initialize $\bT_i^{(0)}$ as $l$ random training examples
  \STATE[2] Initialize $w_i^{(0)}=1/l$ and $q_0\leq 1/2$ such that 
\vspace{-3mm}\[
q_0(1-q_0)=\frac{1}{2n}\min_{i,j} D(\bT_i^{(0)},\bT_j^{(0)}).
\vspace{-3mm}\]
  \STATE [3] E-Step: Compute for each $i=1,...,l$
%\vspace{-2mm}
\[
\begin{split}
f_i(\bx_j)&=q_0^{D(\bx_j,\bT_i^{(0)})}(1-q_0)^{n-D(\bx_j,\bT_i^{(0)})}, j=1,...,m,\\
p_{i}^{(1)}(\bx_j)&=\frac{w_i^{(0)}f_i(\bx_j)}{\sum_{i'} w_{i'}^{(0)} f_{i'}(\bx_j)}, j=1,...,m
\end{split}
%\vspace{-2mm}
\]
 \STATE [4] M-Step: Update, for $i = 1, ..., l$, 
%\vspace{-4mm}
\[
\begin{split}
w_i^{(1)}&=\sum_{j=1}^m p_i^{(1)}(\bx_j)/m\\
\bT_i^{(1)}&=\frac{1}{mw_i^{(1)}}\sum_{j=1}^m p_i^{(1)}(\bx_j)\bx_j
\end{split}
\] %\vspace{-5mm}
\STATE [5] Pruning: Remove all $\bT_i^{(1)}$ with $w_i^{(1)}<w_T=\frac{1}{4l}$
\STATE [6] Pruning: Keep only $k$ templates $\bT_i^{(1)}$ far apart. Let $i = 1, ..., k$ index the remaining $k$ templates. 
\STATE [7] Initialize $w_i^{(1)}=1/k$ and $q_1=q_0$.
 \STATE [8] E-Step: Compute, for $i = 1, ..., k$,  
%\vspace{-3mm}
\[
\begin{split}
f_i(\bx_j)&=q_1^{D(\bx_j,\bT_i^{(1)})}(1-q_1)^{n-D(\bx_j,\bT_i^{(1)})}, j=1,...,m\\
p_{i}^{(2)}(\bx_j)&=\frac{w_i^{(1)}f_i(\bx_j)}{\sum_{i'} w_{i'}^{(1)} f_{i'}(\bx_j)}, j=1,...,m
\end{split}
\]
%\vspace{-4mm}
 \STATE [9] M-Step: Update, for $i = 1, ..., k$,  
 %\vspace{-4mm}
\[
\begin{split}
w_i^{(2)}&=\sum_{j=1}^m p_i^{(2)}(\bx_j)/m,\\
\bT_i^{(2)}&=\frac{1}{mw_i^{(2)}}\sum_{j=1}^m p_i^{(2)}(\bx_j)\bx_j
\end{split}
\]
%\vspace{-4mm}
\end{algorithmic}
\end{algorithm}

After the first iteration, we prune the clusters by a starvation scheme. The pruning process consists of two stages. In the first stage, we remove all the templates $\{\bT_i^{(1)}\}$ whose weights are below a threshold $1/4l$. In the second stage, we keep only $k$ templates that are far apart from each other through an inclusion process. Specifically, we start the inclusion process by randomly picking a template. Then in each subsequent step of the inclusion process, we add a template that is farthest away from the selected templates in terms of the minimum distance between the candidate template and the selected templates. We repeat this step until we get $k$ templates. We let $i = 1, ..., k$ to index the remaining $k$ templates. 

After the pruning process, we run another iteration of EM. The estimated templates from this second round EM are already near optimal as we will show. 

To be more precise, Algorithm 1 describes the two-round EM. In Step 9 the templates $\{\bT_i^{(2)}\}$ are to be converted to binary by rounding to the nearest integer.

%\vspace{-3mm}
\subsection{Notation}
For the convenience of reference, the following summarizes the notation used in this paper: 
%\vspace{-1mm}
\begin{itemize}
\item $n$ is the dimension of Bernoulli templates, which generate examples in $\Omega=\{0,1\}^n$.\vspace{-0mm}
\item $m$ is the number of observations. 
\item $k$ is the true number of clusters. 
\item $q\in (0,1/2)$ is the level of noise\vspace{-0mm}
\item  $B=\displaystyle{\frac{1}{2}(1-2q)\ln \frac{1}{(1-q) (4q+\sqrt{q})}}>0$, 
\item $\displaystyle{
E=\min\left (\frac{1}{4},\frac{c(1-2q)^2/2}{c(1-2q)^2+2(1-q)(q+\sqrt{q})},\frac{3c(1-2q)/4-2q-4\sqrt{6ql/n}}{c(1-2q)+q+\sqrt{q}}\right )}$
%\[E=\displaystyle{\min \left ( \frac{1}{2},\frac{\frac{3}{4}c(1-2q)-2q}{c(1-2q)+2q}\right )}\vspace{-0mm}\]
\item $w_{min}$: the minimum of the mixture weights. \vspace{-0mm}
\item $\bP_i$ is the $i$-th Bernoulli template\vspace{-0mm}
%\item $\bT_i$ is the expected value of the examples from template $\bP_i$, i.e. $\bT_i=E[\bx_i]$ where $\bx_i\sim \bP_i$
\item $S_i$ is the set of examples coming from the template $\bP_i$.\vspace{-0mm}
\item $D(\bP,\bQ)=\sum_{s=1}^{n} |\bP(s)-\bQ(s)|$ is the $\ell_1$ distance between $\bP \in \Omega$ and $\bQ \in \Omega$. 
\item $c_{ij}$ is the separation between the Bernoulli templates, $D(\bP_i,\bP_j)=d_{ij}=nc_{ij}$\vspace{-0mm}
\item $c=\min_{i,j} c_{ij}$\vspace{-0mm}
\item $l$ is the initial number of mixture components $l=\frac{4}{w_{min}}\ln \frac{2}{\delta w_{min}}$. The parameter $\delta$ is the confidence level in Theorem \ref{thm:main}. \vspace{-0mm}
\item $w_T=\frac{1}{4l}$ is the threshold for pruning the clusters learned by the first round. \vspace{-0mm}
\item $C_i$ collects the templates that are initialized from examples in the $i$-th cluster $S_i$ and survive the pruning process after the first round of EM, i.e. \vspace{-0mm}
\[
C_i=\{\bT_{i'}^{(1)}, \bT_{i'}^{(0)}\in S_i, w_{i'}^{(1)}\geq w_T\}\vspace{-0mm}
\]
\end{itemize}
%\vspace{-1mm}

\subsection{Main result}
\begin{theorem} \label{thm:main} Let $m$ examples be generated from a mixture of $k$ Bernoulli templates under Bernoulli noise of level $q$ and  mixing weights $w_i\geq w_{min}$ for all $i$. Let $\epsilon, \delta\in (0,1)$. If  the following conditions hold:
\begin{enumerate}
\item The initial number of clusters is 
\[
l=\displaystyle{\frac{4}{w_{min}}\ln \frac{2}{\delta w_{min}}}.
\]
\item The number of examples is $\displaystyle{m\geq \max(8l, 16\ln n, \frac{8}{w_{min}}\ln \frac{12k}{\delta}})$.
\item The separation is 
\[
\displaystyle{c>\max (\frac{4}{nB}\ln \frac{5n}{\epsilon w_{min}}},\frac{ \max(3(1-2q),2)}{3(1-2q)}(4q+8\sqrt{\frac{6ql}{n}}), \frac{\ln\frac{16l}{\min(6nq,1)}}{nB(1-2q)} ).
\]
\item The dimension is 
\[
\displaystyle{n>\max\left (\frac{3}{\min(c,0.5)E^2}\ln \frac{12 (m+1)^{2}}{\delta},  \frac{6 k}{\delta},\right )}.
\]
\end{enumerate}
Then with probability at least $1-\delta$, the estimated templates after the round 2 of EM satisfy:
\[
D(\bT_i^{(2)},\bP_i)\leq D(\text{mean}(S_i),\bP_i)+\epsilon q
\]
\end{theorem}

The above theorem states that with high probability, the estimated templates from the two-round EM is nearly as accurate as if we knew the memberships of the examples. 

\subsection{Sketch of the proof} The proof follows the steps of the two-round EM. We show that after the initialization, with high probability, the initial templates cover all the clusters and the estimated noise level $q_0$ is close to the true noise level $q$. Then after the first round, the estimated templates are likely to be close to the true templates of the same clusters. After the pruning process, we prove that it is very likely that exactly one template is kept for each cluster. Finally after the second round, the estimated templates are proved to be near optimal. 

\section{Basic facts}

We shall first establish some basic facts about the Bernoulli templates perturbed by Bernoulli noise. They are concerned with the $\ell_1$ distances among templates and their examples. 

\begin{prop} Let $\bP,\bQ\in \Omega$ be Bernoulli templates with noise level $q$. We have: \label{thm:expBern}
\begin{enumerate}
\item If $\bx\sim \bP$ then
\[
\begin{split}
&E[D(\bx,\bP)]=nq, Var[D(\bx,\bP)]=nq(1-q)
\end{split}
\]
\item If $\bx\sim \bP$ and $\by\in \Omega$ then
\[
\begin{split}
E[D(\bx,\by)]&=nq+D(\bP,\by)(1-2q)\\
Var[D(\bx,\by)]&=nq(1-q)
\end{split}
\]
\item If $\bx,\by\sim \bP$ then
\[
\begin{split}
E[D(\bx,\by)]&=2nq(1-q)\\
Var[D(\bx,\by)]&=2nq(1-q)(1-2q+2q^{2})
\end{split}
\]
\item If $\bx\sim \bP,\by\sim \bQ\not =\bP$ then
\[
\begin{split}
E[D(\bx,\by)]&=2nq(1-q)+D(\bP,\bQ)(1-2q)^{2}\\
Var[D(\bx,\by)]&=2nq(1-q)(1-2q+2q^{2})
\end{split}
\]
\end{enumerate}
\end{prop}

\begin{prop} Let $\bP,\bQ\in \Omega$ be  Bernoulli templates with noise level $q$.  We have: \label{thm:devBern}
\begin{enumerate}
\item[a)] If $\bx\sim \bP$ and $\lambda \geq 1$ then
\[
\bP(D(\bx,\bP)>\lambda nq )\leq e^{-nq(\lambda-1)^{2}/3}
\]
\item[b)] If $\bx\sim \bP$ and $\epsilon \in (0,1)$ then
\[
\begin{split}
&\bP(|D(\bx,\bP)-nq|>\epsilon n\sqrt{q} )\leq 2e^{-n\epsilon^{2}/3}
\end{split}
\]
\item[c)] If $\bx\sim \bP, \by\sim \bQ$  and 
\[\nu(\bP,\bQ)=2nq(1-q)+D(\bP,\bQ)(1-2q)^{2}\]
then for any $\epsilon \in(0,1)$
\[
\bP(|D(\bx,\by)-\nu(\bP,\bQ)|>\epsilon \nu(\bP,\bQ) )\leq 2e^{-\nu(\bP,\bQ) \epsilon^{2}/3}
\]
\end{enumerate}
\end{prop}

Prop. \ref{thm:devBern} states that the $\ell_1$ distance between an example and its template is concentrated around $nq$, while the distance between two examples from two different templates is concentrated around $\nu(\bP, \bQ)$. This leads to the following proposition. 

\begin{prop} Draw $m$ samples from a c-separated mixture of $k$ Bernoulli templates with mixing weights at least $w_{min}$. \label{prop:eps}
Let $\epsilon_0>0$. Then with probability at least $1-m^{2}e^{-2n(1-q) \epsilon_0^{2}/3}
-m^{2}e^{-n\min(c,0.5) \epsilon_0^{2}/3}-2me^{-n\epsilon_0^{2}/3}-2me^{-n\min(c,0.5)\epsilon_0^{2}/3}-ke^{-mw_{min}/8}$ 
\begin{enumerate}
\item[a)] For any $\bx,\by\in S_i$ we have
\[
D(\bx,\by)=2n(1-q)(q\pm \epsilon_0 \sqrt{q} )
\]
\item[b)] For any $\bx\in S_i, \by\in S_j$, $i \neq j$,  we have
\[
D(\bx,\by)=n(2q(1-q)+c_{ij}(1-2q)^{2})(1\pm \epsilon_0)
\]
\item[c)] For any $\bx\in S_i$  we have
\[
\begin{split}
D(\bx,\bP_i)&=n(q\pm \epsilon_0 \sqrt{q})\\
D(\bx,\bP_j)&=n(q+c_{ij}(1-2q))(1\pm \epsilon_0)
\end{split}
\]
\item[d)] Each $|S_i|\geq \frac{1}{2}mw_i$.
\end{enumerate}
\end{prop}
Here we employ the notation that $a = b \pm \epsilon$ means $a \in (b-\epsilon, b+\epsilon)$.

\begin{lem} Let  $Z_i=\frac{1}{m}\sum_{j=1}^m B_{ij}$ where $B_{ij}$ are Bernoulli random variables with $E[B_{ij}]=q$. Then \label{lem:absZq}
\[
\bP(\sum_{i=1}^n Z_i -nq >\lambda)<\exp(-\frac{m \lambda^{2}}{3nq})
\]
\end{lem}

\begin{prop}{\bf (Average of subsets)} Draw a set $S_1$ of $m$ examples randomly from template $\bP\in \{0,1\}^n$ with noise level $q<1/2$. Then with probability at least $1-\delta$ for any subset of size at least $t\geq n$ there is no subset of $S_1$ of size at least $t$ whose average $\mu$ has 
\label{prop:avg1}
\[
D(\mu,\bP)\geq nq+ \sqrt{3nq\left ( \ln \frac{me}{t}+ \frac{1}{t}\ln  \frac{1}{\delta}\right )}
\]
%where $A=e^{-(0.5-q)^{2}/3q}$.
\end{prop}

Prop. \ref{prop:avg1} states that the sample average is unlikely to deviate too far from $\bP$. 

\begin{prop}{\bf (Weighted averages)} For any finite set of points $S\subset \{0,1\}^n$ and weights $w_\bx\in [0,1],\bx\in S$ there exists a subset $T\subset S$ such that \label{prop:wtavg}
\begin{enumerate}
\item $|T|= \lfloor \sum_{\bx\in S} w_\bx\rfloor$
\item $D(\mu_T,\bP)\geq D(\mu_w,\bP)$ where
\[ \mu_T=\frac{1}{|T|}\sum_{\bx\in T} \bx  \text{ and } \mu_w=\frac{\sum_{\bx\in S} w_\bx \bx}{\sum_{\bx\in S} w_\bx}.\] 
\end{enumerate}
\end{prop}

Prop. \ref{prop:wtavg} states that the weighted average can be bounded by unweighted average. This result is needed because the templates are estimated as the weighted averages in both rounds of the EM algorithm and from Prop. \ref{prop:avg1} and \ref{prop:wtavg} we can bound on the distance to the template.

\section{Key steps of the proof}

In this section we state the results that hold for the estimated template parameters after each EM iteration. We assume that the following technical conditions hold 
\begin{enumerate}
\item[C1:] $nc>\displaystyle{\frac{1}{B(1-2q)}\ln\frac{16l}{\min(6nq,1)}}$ 
%or equivalently $e^{-ncB(1-2q)/2}< \frac{1}{8l}c =\frac{w_T}{2}c $
\item[C2:] $m>\max(16\ln n,8l)$
\item[C3:] $\displaystyle{c>\max(1,\frac{2}{3(1-2q)}) }(4q+8\sqrt{6ql/n})$
\end{enumerate}
These conditions are a subset of the conditions of Theorem \ref{thm:main} that don't depend on $w_{min}$ and $\delta$. They will be referred to in the proofs of the statements of this section.

We also assume that $\epsilon_0\leq E$ where condition C3 guarantees that $E>0$.
Observe that condition $C3$ imposes an upper bound on the noise level $q$ since $c<1$. In our experiments this upper bound was between $0.2$ and $0.3$.

\subsection{Initialization}
This section analyzes the initial estimates for the parameters before the first round of EM.

\begin{prop} With probability at least $1-k(l+1)e^{-lw_{min}}-ke^{lw_{min}/4}$  we have \label{prop:em0}
\begin{enumerate}
\item For each true template $\bP_i$, the number of  $\bT_j^{(0)}$ coming from $\bP_i$ is at least 2. 
\item For each true template $\bP_i$, the number of  $\bT_j^{(0)}$ coming from $\bP_i$ is at most $\frac{15}{8}lw_i$
\item The noise estimate satisfies
\[
q_0(1-q_0)=(1-q)(q\pm \epsilon_0 \sqrt{q}).%, \; q_0=q(1\pm \sqrt{ \epsilon})
\]
\end{enumerate}
\end{prop}

By initializing from more templates than the actual number of clusters, there is a high probability that the estimated templates cover all the clusters. 

\subsection{First Round of EM}
\begin{prop} Suppose $\bT_{i'}^{(0)}\in S_i$ and $\bT_{j'}^{(0)}\in S_j$, $i \neq j$. In the cases when the conclusions of Proposition \ref{prop:eps} hold,  for any $\bx\in S_i$ the ratio between the probabilities $p_i$ and $p_j$ is
\label{prop:probs} 
\[
\frac{p_{i'}^{(1)}(\bx)}{p_{j'}^{(1)}(\bx)}\geq \exp(nc_{ij}B(1-2q))
\]
\end{prop}

Prop. \ref{prop:probs} states that the first round of EM will likely give higher weights to the templates representing the correct cluster than to a wrong cluster. 

\begin{prop}  In the cases when the conclusions of Proposition \ref{prop:eps}  hold, any non-starved  estimate $\bT_{i'}^{(1)}\in C_i$ satisfies with probability $1-1/n$ \label{prop:em1}
\[
D(\bT_{i'}^{(1)},\bP_i)\leq  nq+\sqrt{6nql}
\]
\end{prop}

So the estimated template of a cluster is very likely to be close to the true template of this cluster. 

\subsection{Pruning}

We prove that with high probability the pruning step will keep exactly one template from each cluster.

\begin{prop}In the cases when Propositions \ref{prop:eps}, \ref{prop:em0} and \ref{prop:em1} hold, the set $C_i$ obeys the following properties:\label{prop:pruning}
\begin{enumerate}
\item[a)] Each $C_i$ is non-empty
\item[b)] There exists $\tau\in \RR$ such that for any $\bx\in C_i$ and $\by,\bz\in C_j, j\not = i$ we have $D(\by,\bz)\leq \tau$ and $D(\bx,\by)>\tau$.
\item[c)] The pruning procedure finds exactly one member of each $C_i$.
\end{enumerate}
\end{prop}

\subsection{Second Round of EM}
We permute the obtained templates $\bT_i^{(1)}$ so that  $\bT_i^{(1)}\in S_i$.

\begin{prop} Suppose $\bT_i^{(1)}\in S_i$ and $\bT_j^{(1)}\in S_j$, $i \neq j$.  In the cases when Propositions \ref{prop:eps}, \ref{prop:em0} and \ref{prop:em1} hold, for any $\bx\in S_i$ the ratio between the probabilities $p_i$ and $p_j$ is
\label{prop:probs2} 
\[
\frac{p_i^{(2)}(\bx)}{p_j^{(2)}(\bx)}\geq \exp(\frac{1}{4}nc_{ij}(1-2q)\ln \frac{1}{6\sqrt{q}})=\exp(n c_{ij} B/2)
\]
\end{prop}

\begin{theorem} \label{thm:em2}
Suppose that $l>k$, $w_i>w_{min}$ for all $i$ and that conditions $C1-C3$ hold. Then with probability at least $1-m^{2}e^{-2n(1-q) \epsilon_0^{2}/3}
-m^{2}e^{-n\min(c,0.5) \epsilon_0^{2}/3}-2me^{-n\epsilon_0^{2}/3}-2me^{-n\min(c,0.5)\epsilon_0^{2}/3}-ke^{-mw_{min}/8}-k(l+1)e^{-lw_{min}}-ke^{lw_{min}/12}-k/n$, the estimated templates after the round 2 of EM satisfy:
\[
D(\bT_i^{(2)},\bP_i)\leq D(\text{mean}(S_i),\bP_i)+\frac{5}{w_{min}}e^{-ncB/4}nq
\]
\end{theorem}

We are now ready to prove Theorem \ref{thm:main}. 

{\em Proof of Theorem \ref{thm:main}}.

From  $l=\displaystyle{\frac{4}{w_{min}}\ln \frac{2}{\delta w_{min}}}$, we get $ke^{-lw_{min}/4}=k\delta w_{min}/2\leq \delta/2$. 
Also
\[
\begin{split}
k(l+1)e^{-lw_{min}}&<2kle^{-lw_{min}/12}e^{-11lw_{min}/12}\\
&\leq \frac{\delta}{2} 2l \frac{\delta^{11}w_{min}^{11}}{2^{11}}=lw_{min}\delta^{11}w_{min}^{10}\delta/2^{10}
\end{split}
\]
But 
\[
lw_{min}=12\ln  \frac{2}{\delta w_{min}}\leq \frac{24}{\delta w_{min}}
\]
so
\[
k(l+1)e^{-lw_{min}}<24 \delta^{10}w_{min}^{9}\delta/2^{10}<\delta/12
\]

Take $\epsilon_0=E>0$ (because of C3). From the dimension condition 
\[
\displaystyle{n>\frac{3}{\min(c,0.5)E^{2}}\ln \frac{12 (m+1)^{2}}{\delta}}
\]
 we get $(m+1)^{2} e^{-n \min(c,0.5)\epsilon_0^{2}/3}\leq \delta/12$,
so
\begin{eqnarray*}
m^{2}e^{-2n(1-q) \epsilon_0^{2}/3}+2me^{-n \epsilon_0^{2}/3}+
m^{2}e^{-n\min(c,0.5) \epsilon_0^{2}/3}+2me^{-n\min(c,0.5)\epsilon_0^{2}/3}\\
\leq 2(m+1)^{2} e^{-n \min(c,0.5)\epsilon_0^{2}/3}\leq \delta/6.
\end{eqnarray*}

From the dimension condition $\displaystyle{n> {6 k}/{\delta}}$ we get $k /n<\delta/6$.

From the condition on the number of examples, we get $ke^{-mw_{min}/8}<\delta/12$.

From Theorem \ref{thm:em2}, putting all of the above inequalities together and taking $\displaystyle{nc>\frac{4}{B}\ln \frac{5n}{\epsilon w_{min}}}$, we obtain that Theorem \ref{thm:main} holds with probability at least $1-\delta$. $\Box$

\section{Experiments}

This section illustrates the theoretical results obtained in the previous sections by a simulation study as well as experiments on synthetic image sketches and real images.

\subsection{Simulation study}

In this section we conduct experiments showing that indeed, the true templates are found with high probability when the conditions of Theorem \ref{thm:main} hold.

We will work with a mixture of two templates, $\bP_1={\bf 0}$ and $\bP_2=(1,1,...,1,0,0,..,0)$ where the number of 1's is $\lfloor cn \rfloor$, to obtain a desired separation $c\in [0,1]$ in dimension $n$. We experiment with standard EM for 2, 10 and 20 iterations. The standard EM starts from $k$ clusters, instead of $l$ clusters followed by pruning as in the two-step EM. For the standard EM we also assumed the noise level $q$ is a known parameter. All results are obtained from 100 runs. 
\begin{figure}[htb]
\centering
\includegraphics[width=6.cm]{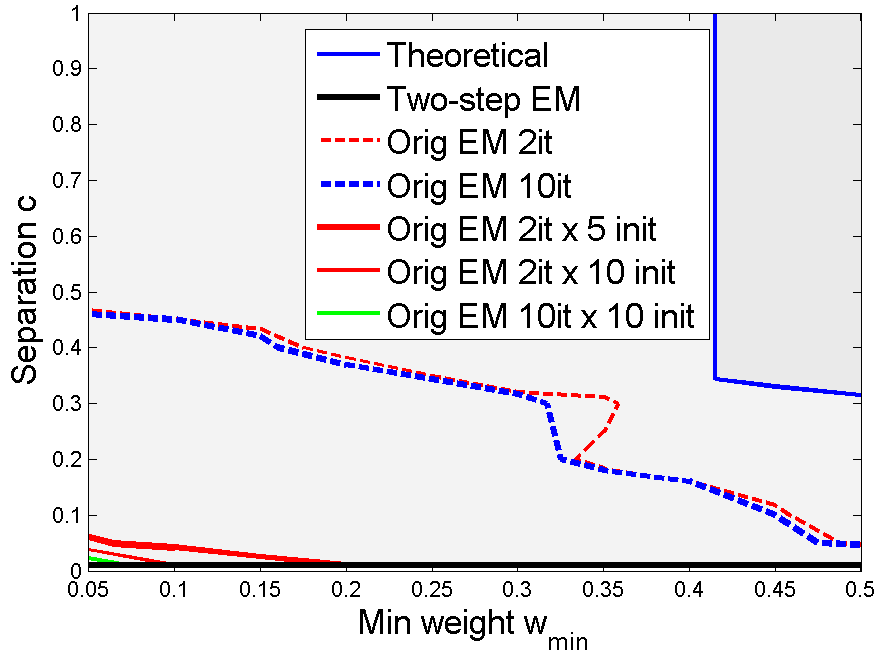}
\hspace{-1.5mm}\includegraphics[width=6.cm]{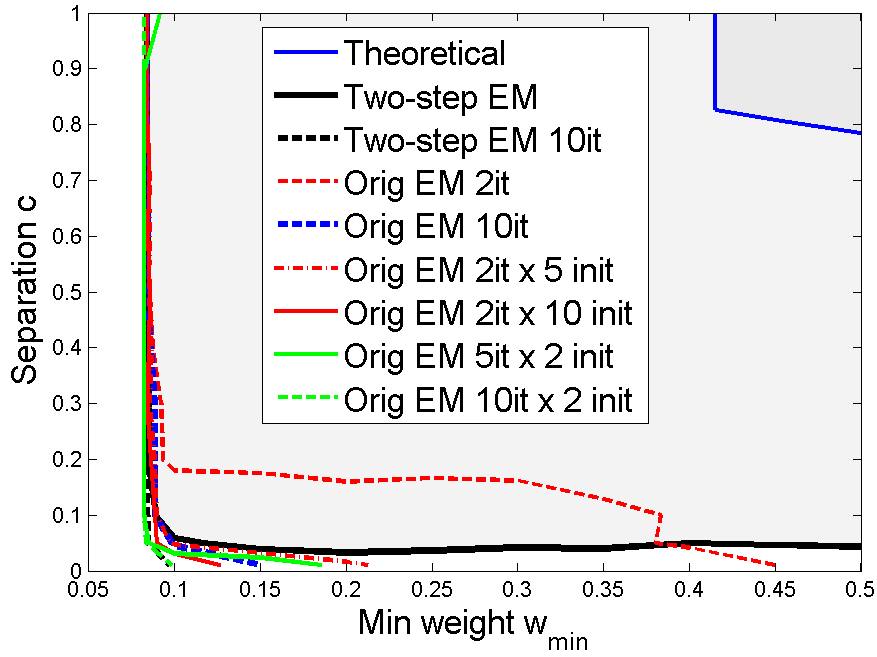}
\vskip -3.mm
\caption{Domains where the two-round EM and the standard EM find the $k=2$ binary templates correctly $90\%$ of the time when $m = 300$. The first plot is for $q = .01$, with $n = 2,000$,  and the second plot is for $q = .1$ with $n = 10,000$. Also shown is the domain theoretically guaranteed by Theorem \ref{thm:main}. Each domain is above and to the right of the corresponding curve. }
\label{fig:plotdomainw}
\vspace{-4mm}
\end{figure}

\begin{figure}[ht]
\centering
\hspace{-1.mm}\includegraphics[width=6.cm]{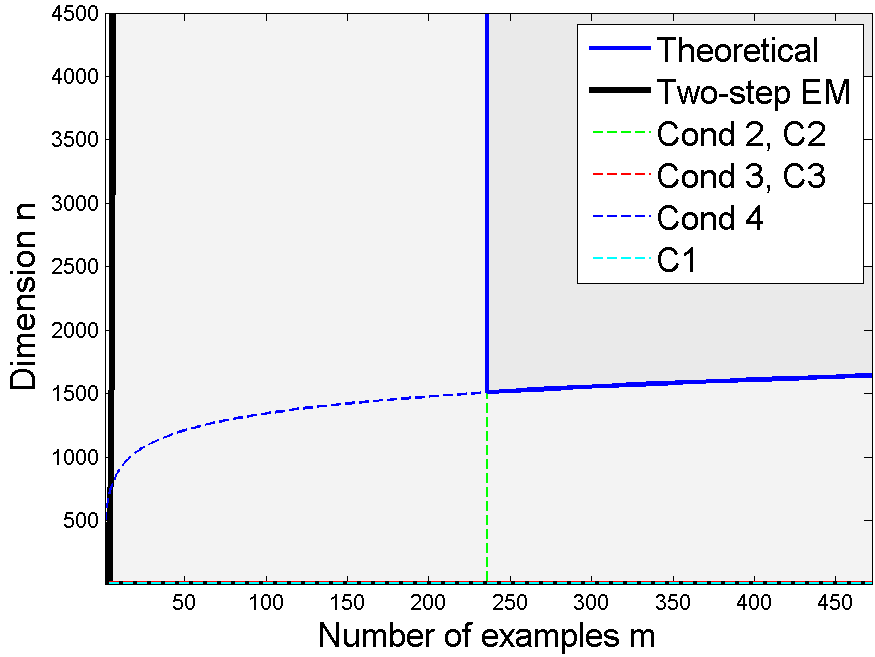}
\hspace{-1.mm}\includegraphics[width=6.cm]{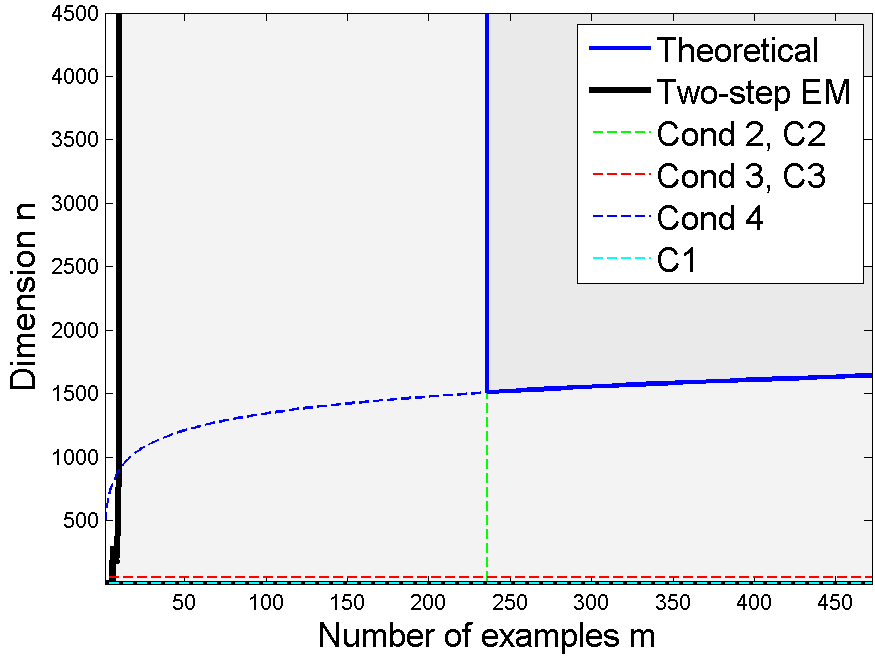}
\hspace{-1.mm}\includegraphics[width=6.cm]{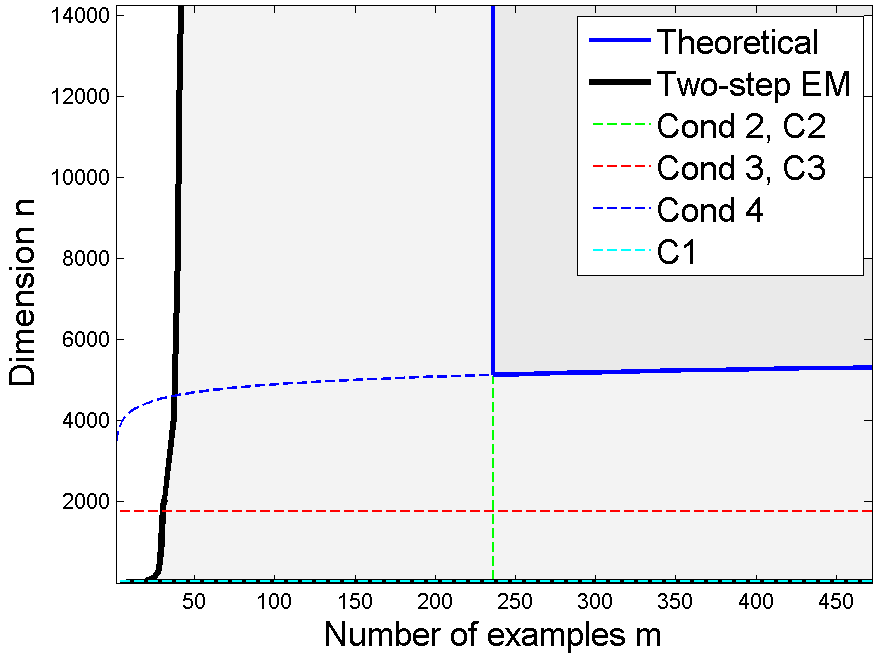}
\hspace{-0.9mm}\includegraphics[width=6.cm]{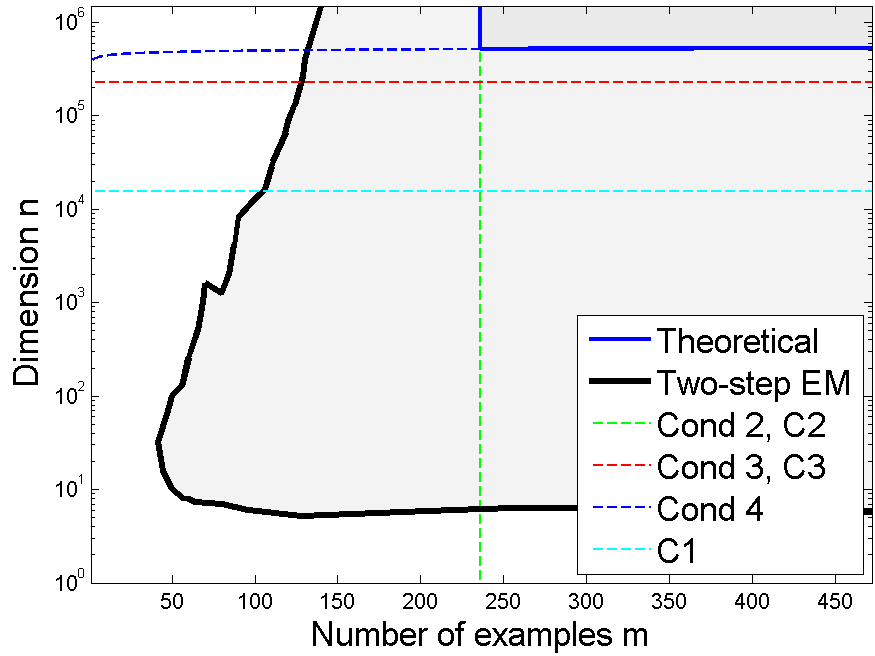}
\vskip -3.mm
\caption{Theoretical and practical domains of validity of the two-step EM algorithm for four noise levels. From left to right are noise levels: $q=0.0001, q=0.01$ (top) and $q=0.1,q=0.2$ (bottom). In these examples $c=1,k=2,w_{min}=0.5,\delta=\epsilon=0.1$. Each domain is above and to the right of the corresponding curve. }
\label{fig:plotdomain}
\vspace{-3mm}
\end{figure}

Fig. \ref{fig:plotdomainw} and  \ref{fig:plotdomain} show the domains where the two-step EM and the standard EM  find the templates  $\bP_1,\bP_2$ with $90\%$ probability, thus $\delta=0.1$. 

In the two plots of Fig. \ref{fig:plotdomainw}, the horizontal axis is the minimum weight $w_{\rm min}$, and the vertical axis is the separation $c$. The domain for each algorithm is the region above and to the right of the corresponding curve. Two version of the two step EM algorithm were evaluated: the two-step EM, and 10-step version that does 9 EM steps after the pruning step. Five version of the original EM were evaluated, with 2 or 10 iterations, and 1, 5 or 10 random initializations (and selecting from the 5 or 10 obtained results the largest likelihood one as the final result).
The first plot  is obtained at the noise level $q = .01$, while the second plot is for the noise level $q = .1$. We take the number of observations $m = 300$. For the first plot the dimension is $n= 2,000$ and for the second plot, $n = 10,000$. One can see that for low noise, the two-step EM works better than the original EM. Also displayed is the domain where the conditions of our theorem are satisfied.

In the four plots from Fig.  \ref{fig:plotdomain} the horizontal axis is the number $m$ of observations  and the vertical axis is the dimension $n$. 
The four plots show the domain where the two-step EM algorithm finds the templates  $\bP_1,\bP_2$ with $90\%$ probability for the levels of noise $q\in \{0.0001,0.01,0.1,0.2\}$. The curves corresponding to conditions 2-4 of Theorem \ref{thm:main} and the technical conditions C1-C3 are also displayed, as well as the domain where all conditions of our theorem are satisfied.

From the experiments we observe that the domain where the templates are found with high probability is larger than the domain where the conditions of  Theorem \ref{thm:main} hold. The largest discrepancy is in the dimensionality conditions, where the gap between theory and experiments is considerable. This gap could be substantially decreased if tighter bounds could be obtained for Prop \ref{prop:avg1} and consequently for Prop \ref{prop:em1} and Theorem \ref{thm:em2}.

\begin{figure}[ht]
\centering
\includegraphics[width=1.7cm]{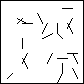}
\includegraphics[width=1.7cm]{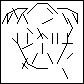}
\includegraphics[width=1.7cm]{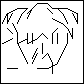}
\includegraphics[width=1.7cm]{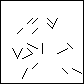}
\includegraphics[width=1.7cm]{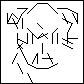}\\ \vskip 1mm
\includegraphics[width=1.7cm]{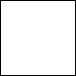}
\includegraphics[width=1.7cm]{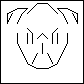}
\vskip -3mm
\caption{Top row: Examples of training images. Bottom row: the Bernoulli templates used to generate the training images.}
\label{fig:dog_examples}
\vspace{-5mm}
\end{figure}
%\vskip -3mm
\begin{figure}[htb]
\centering
\hspace{-1mm}\includegraphics[width=6.cm]{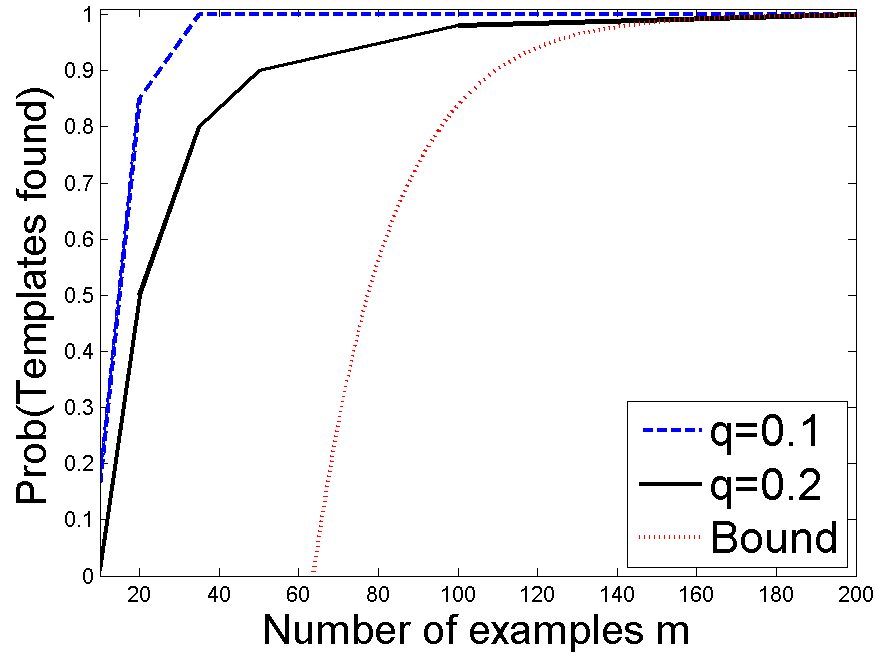}
\hspace{-1mm}\includegraphics[width=6.cm]{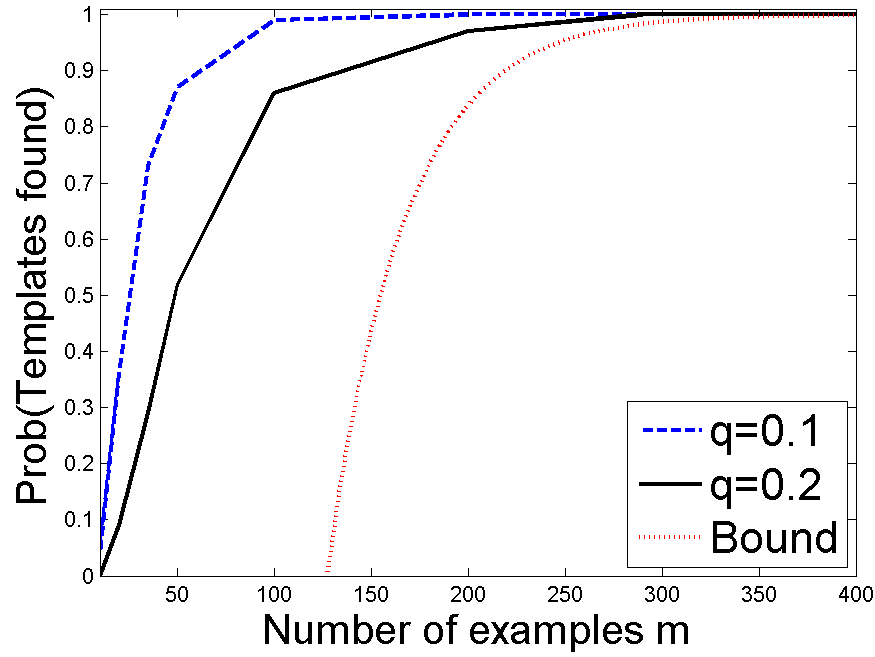}
\vskip -3.mm
\caption{Success rates vs. number of training examples for learning from a mixture of two templates with the two-round EM algorithm for two levels of noise $q\in \{0.1,0.2\}$ and two mixture weights $w_{min}=0.4$ (left) and $w_{min}=0.2$ (right).}
\label{fig:plotdelta_em_and}
\vspace{-6mm}
\end{figure}
\subsection{Experiments on synthetic image sketches}

In this experiment we work with a mixture of two Bernoulli templates, shown in the bottom row of Fig. \ref{fig:dog_examples}, in a space of dimension $n=9\times 9\times 18=1458$. By perturbing the entries with Bernoulli noise of level $q$ we obtain images such as those shown in the top row of Fig. \ref{fig:dog_examples}.

Fig. \ref{fig:plotdelta_em_and} shows the success rate of finding the two templates exactly using the two-round EM algorithm vs. the number of training examples. The experiments are run for two levels of noise  $q\in \{.1,.2\}$ and two mixture weights $w_{min}\in \{.2,.4\}$.

Also shown is the bound $1-\delta>1-12ke^{-mw_{min}/8}$ from condition 2 of Theorem \ref{thm:main}. 

The separation between the two templates is quite small $c=.02$, because the two templates share a lot of zero components. So the separation conditions fail in this case. Since we are not in the conditions of the Theorem 1, the bound on the training examples is not expected to hold. We may achieve a better bound if we reduce the dimension $n$ while increasing $c$ by selecting those features that differentiate the templates. In any case, we see that in the given scenarios the two templates can be recovered with 100\% certainty with the two-round EM given sufficiently many examples. So Theorem \ref{thm:main} might hold under milder assumptions than ours.

\subsection{Experiments on real images}

\begin{figure*}[htb]
\centering
\includegraphics[width=12.cm]{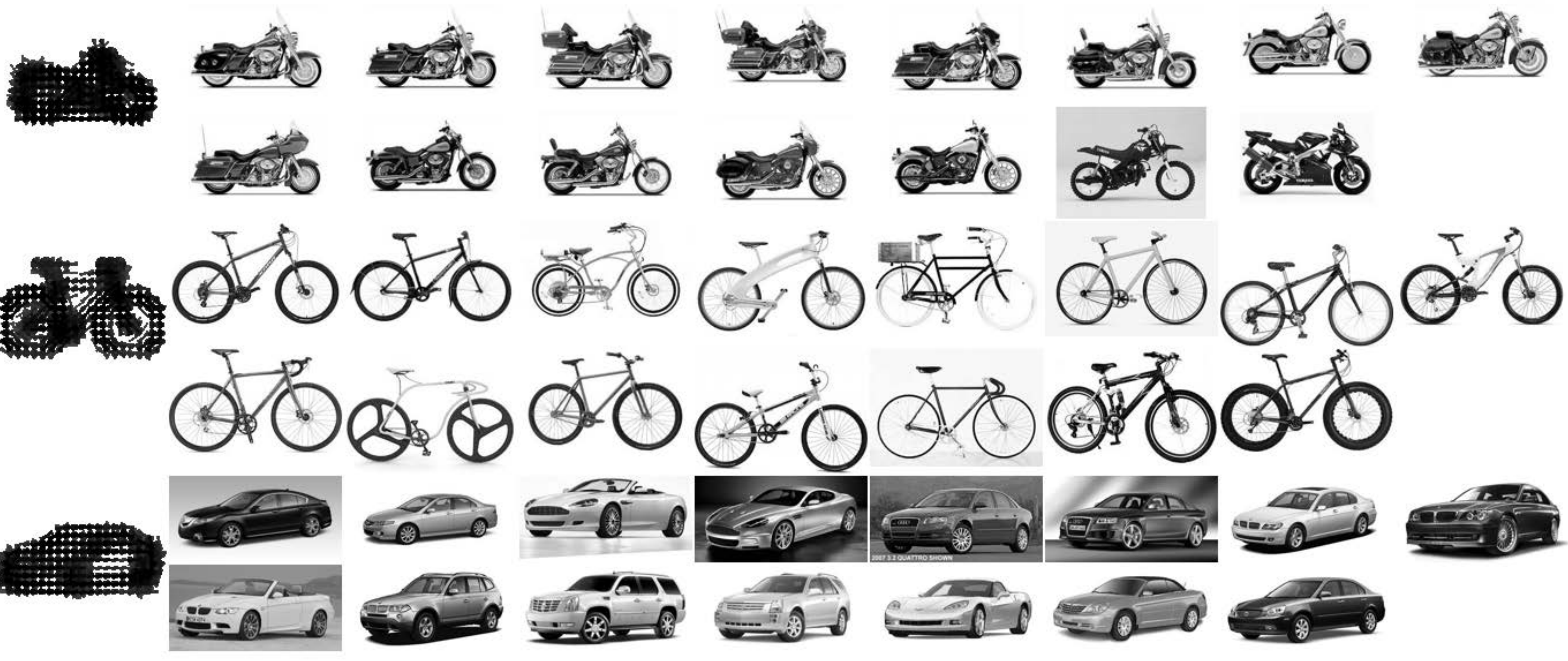}
\vskip -4mm
\caption{Clustering motorcycles, bicycles and cars by the two-round EM algorithm. In each row, the first plot displays the learned template and the rest of the plots show some of the examples in the corresponding cluster. There are 15 images in each cluster. }
\label{fig:CarBicycleMotorcycle}
\vspace{-4mm}
\end{figure*}
%\begin{figure*}[h]
%\centering
%\includegraphics[width=12.5cm]{face.jpg}
%%\vskip -3.mm
%\caption{Clustering wolves, deer, cats and rabbits. In each row, the first plot displays the learned template and the rest of the plots show some of the examples in the corresponding cluster. There are 15 images in each cluster. 
%\label{fig:real1}}
%%\vspace{-6mm}
%\end{figure*}

%\vskip -4.mm

%=============================================================================
%\begin{table*}
%\begin{center}
%    \begin{tabular}{ | l | l | l | l | l | }
%    \hline    
%   	\multicolumn{2}{|l|}{} & Cond. Purity & Cond. Entropy & Num$_{\text{perfect}}$ /100 \\ \hline
%   	\multicolumn{2}{|l|}{Two-round EM} &  \textbf{0.944}$\pm$ 0.108 & \textbf{0.106}$\pm$ 0.183 & \textbf{61} \\ \hline
%   	\multirow{4}{*}{Original EM} &  maxIter=2 & 0.851$\pm$0.157 & 0.247$\pm$ 0.231 & 25 \\ \cline{2-5}
%   	&  maxIter=10 & 0.886$\pm$0.155 & 0.170$\pm$0.211 & 47 \\ \cline{2-5}
%   	&  maxIter=20 & 0.886$\pm$0.155 & 0.170$\pm$0.211 & 47 \\ \cline{2-5}
%   	&  maxIter=100 & 0.886$\pm$0.155 & 0.170$\pm$0.211 & 47 \\ \hline
%    \end{tabular}
%\end{center}
%\caption{Performance comparison between our two-round EM algorithm and the original EM algorithm for clustering motorcycles, bicycles and cars. In the table, Num$_{\text{perfect}}$ means the number of runs (out of the total 100 runs) that recover the underlying clusters perfectly. }\label{table:CarBicycleMotorcycle} 
%\end{table*}

%=============================================================================
\begin{table*}[ht]
\begin{center}
    \begin{tabular}{ | l | l | l | l | l | l | }
    \hline    
   	Method & \parbox{1.2cm}{\#Round ($\times$\#Init)} & Cond. Purity & Cond. Entropy & Log-Likelihood & $N$ \\ \hline
   	\multirow{2}{*}{\parbox{1.5cm}{Tow-round EM}} 
   	& 2 ($\times$1)  & 0.9402 $\pm$ 0.1124 & 0.1098 $\pm$ 0.1862 & -110625.6 $\pm$ 7914.1 & 61 \\ \cline{2-6}
    & 10 ($\times$1) & 0.9822 $\pm$ 0.0653 & 0.0351 $\pm$ 0.0937 & -108460.3 $\pm$ 6491.8 & 76 \\ \hline
   	\multirow{12}{*}{\parbox{1.5cm}{Original EM}} 
   	&  2 ($\times$1) & 0.8511 $\pm$ 0.1500 & 0.2464 $\pm$ 0.2193 & -115852.1 $\pm$ 11079.3 & 27 \\ \cline{2-6}
   	&  10 ($\times$1) & 0.9004 $\pm$ 0.1464 & 0.1555 $\pm$ 0.2000 & -113476.2 $\pm$ 10889.9 & 43 \\ \cline{2-6}
   	&  20 ($\times$1) & 0.8722 $\pm$ 0.1572 & 0.1917 $\pm$ 0.2144 & -115083.1 $\pm$ 10828.8 & 40 \\ \cline{2-6}
   	&  100 ($\times$1) & 0.9051 $\pm$ 0.1460 & 0.1447 $\pm$ 0.2006 & -113205.1 $\pm$ 10809.3 & 51 \\ \cline{2-6} \cline{2-6}

   	&  2 ($\times$5) & 0.9911 $\pm$ 0.0295 & 0.0260 $\pm$ 0.0599 & -106268.2 $\pm$ 5448.4 & 77 \\ \cline{2-6}
   	&  10 ($\times$5) & 0.9987 $\pm$ 0.0053 & 0.0050 $\pm$ 0.0198 & -106067.0 $\pm$ 5122.5 & 94 \\ \cline{2-6}
   	&  20 ($\times$5) & 0.9956 $\pm$ 0.0336 & 0.0088 $\pm$ 0.0493 & -106249.3 $\pm$ 5540.3 & 94 \\ \cline{2-6}
   	&  100 ($\times$5) & 0.9996 $\pm$ 0.0031 & 0.0017 $\pm$ 0.0117 & -106045.7 $\pm$ 5106.5 & 98 \\ \cline{2-6} \cline{2-6}

   	&  2 ($\times$10) & 0.9991 $\pm$ 0.0054 & 0.0030 $\pm$ 0.0179 & -107549.1 $\pm$ 5366.8 & 97 \\ \cline{2-6}
   	&  10 ($\times$10) & 1.0000 $\pm$ 0.0000 & 0.0000 $\pm$ 0.0000 & -107534.8 $\pm$ 5377.5 & 100 \\ \cline{2-6}
   	&  20 ($\times$10) & 1.0000 $\pm$ 0.0000 & 0.0000 $\pm$ 0.0000 & -107534.8 $\pm$ 5377.5 & 100 \\ \cline{2-6}
   	&  100 ($\times$10) & 0.9998 $\pm$ 0.0022 & 0.0008 $\pm$ 0.0083 & -107541.0 $\pm$ 5390.3 & 99 \\ \hline 
   	
    \end{tabular}
\end{center}
\vskip -1mm
\caption{Comparison of the two-step EM algorithm and the original EM for clustering motorcycles, bicycles and cars. Shown are the mean$\pm$std of the conditional purity, conditional entropy and log-likelihood. $\#$Round is the number of steps of an EM algorithm, and $\#$Init the number of random initializations used to select the best result (in terms of the log-likelihood). $N$ is the number of runs (out of  100 total runs) that recover the clusters perfectly.  }\label{table:CarBicycleMotorcycle} 
\vspace{-7mm}
\end{table*}

We also performed experiments on real images. Each image is first convolved with Gabor filters tuned to 16 orientations. Then the image domain is partitioned into equal sized squared cells (the size ranges from $5 \times 5$ pixels to $7 \times 7$ pixels). Within each cell, at each orientation, we pool the maximum of the absolute values of the filter responses. If the maximum is above a threshold, we declare that there is a sketch within this cell at this orientation. Thus each cell produces a binary vector of 16 components. We then concatenate the binary vectors of all the cells into a large binary vector. So each image is transformed into a binary vector. 
\begin{figure*}[h]
\centering
\includegraphics[width=12.cm]{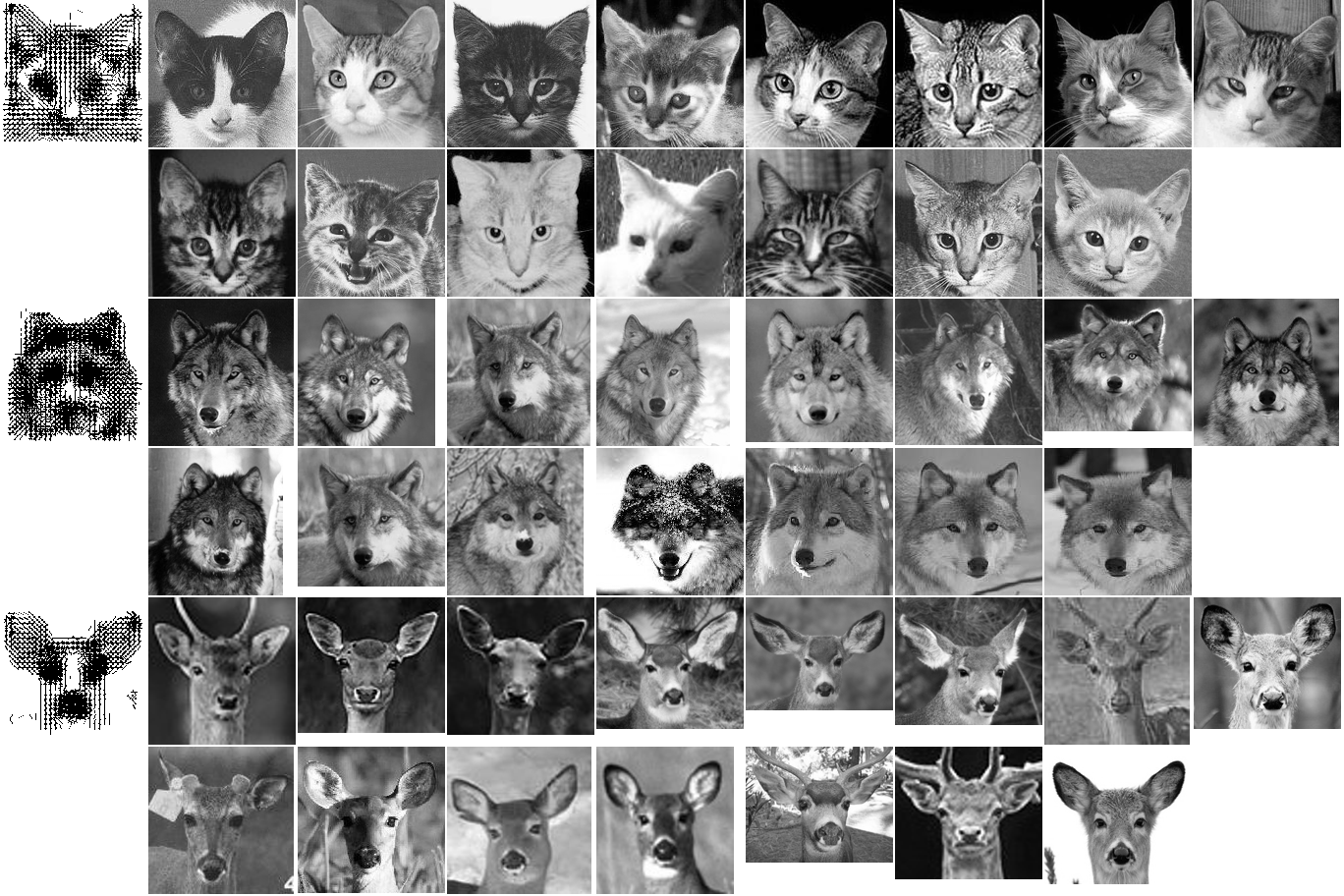}
\vskip -3.mm
\caption{Clustering cats, wolves and deers by the two-round EM algorithm. In each row, the first plot displays the learned template and the rest of the plots show some of the examples in the corresponding cluster. There are 15 images in each cluster. }
\label{fig:CatDeerWolf}
\vspace{-4mm}
\end{figure*}

%=============================================================================
%\begin{table*}
%\begin{center}
%    \begin{tabular}{ | l | l | l | l | l | }
%    \hline    
%   	\multicolumn{2}{|l|}{} & Cond. Purity & Cond. Entropy & Num$_{\text{perfect}}$ /100 \\ \hline
%   	\multicolumn{2}{|l|}{Two-round EM} &  \textbf{0.846}$\pm$ 0.139 & \textbf{0.319}$\pm$ 0.234 & \textbf{12} \\ \hline
%   	\multirow{4}{*}{Original EM} &  maxIter=2 & 0.704$\pm$0.139 & 0.581$\pm$ 0.245 & 2 \\ \cline{2-5}
%   	&  maxIter=10 & 0.744$\pm$0.152 & 0.494$\pm$0.271 & 8 \\ \cline{2-5}
%   	&  maxIter=20 & 0.744$\pm$0.152 & 0.494$\pm$0.271 & 8 \\ \cline{2-5}
%   	&  maxIter=100 & 0.744$\pm$0.152 & 0.494$\pm$0.271 & 8\\ \hline
%    \end{tabular}
%\end{center}
%\caption{Performance comparison between our two-round EM algorithm and the original EM algorithm for clustering cats, wolves and deers. In the table, Num$_{\text{perfect}}$ means the number of runs (out of the total 100 runs) which recover the underlying clusters perfectly. }\label{table:CatDeerWolf} 
%\end{table*}

%=============================================================================
\begin{table*}[ht]
\begin{center}
    \begin{tabular}{ | l | l | l | l | l | l | }
    \hline    
   	Method & \parbox{1.5cm}{\#Round ($\times$\#Init)} & Cond. Purity & Cond. Entropy & Log-Likelihood & $N$ \\ \hline
   	\multirow{2}{*}{\parbox{1.5cm}{Tow-round EM}} 
   	& 2 ($\times$1)  & 0.8918 $\pm$ 0.1268 & 0.2450 $\pm$ 0.2230 & -398033.9 $\pm$ 14285.5 & 17 \\ \cline{2-6}
    & 10 ($\times$1) & 0.9222 $\pm$ 0.1169 & 0.1705 $\pm$ 0.2016 & -396428.5 $\pm$ 13583.5 & 32 \\ \hline
   	\multirow{12}{*}{\parbox{1.5cm}{Original EM}} 
   	&  2 ($\times$1) & 0.6853 $\pm$ 0.1472 & 0.6300 $\pm$ 0.2429 & -408941.0 $\pm$ 16116.3 & 2 \\ \cline{2-6}
   	&  10 ($\times$1) & 0.7342 $\pm$ 0.1386 & 0.5151 $\pm$ 0.2341 & -406125.6 $\pm$ 13973.7 & 2 \\ \cline{2-6}
   	&  20 ($\times$1) & 0.7489 $\pm$ 0.1484 & 0.4876 $\pm$ 0.2534 & -405384.8 $\pm$ 15374.4 & 5 \\ \cline{2-6}
   	&  100 ($\times$1) & 0.7493 $\pm$ 0.1426 & 0.4788 $\pm$ 0.2497 & -405643.4 $\pm$ 16290.6 & 6 \\ \cline{2-6} \cline{2-6}

   	&  2 ($\times$5) & 0.8587 $\pm$ 0.1039 & 0.3459 $\pm$ 0.1985 & -562506.3 $\pm$ 45380.3 & 6 \\ \cline{2-6}
   	&  10 ($\times$5) & 0.9276 $\pm$ 0.0881 & 0.1895 $\pm$ 0.1795 & -556137.0 $\pm$ 45076.8 & 19 \\ \cline{2-6}
   	&  20 ($\times$5) & 0.9027 $\pm$ 0.0985 & 0.2373 $\pm$ 0.1861 & -558410.6 $\pm$ 45627.2 & 11 \\ \cline{2-6}
   	&  100 ($\times$5) & 0.9287 $\pm$ 0.0835 & 0.1834 $\pm$ 0.1656 & -555971.3 $\pm$ 44003.9 & 20 \\ \cline{2-6} \cline{2-6}

   	&  2 ($\times$10) & 0.9242 $\pm$ 0.0731 & 0.2114 $\pm$ 0.1688 & -395855.2 $\pm$ 11914.2 & 15 \\ \cline{2-6}
   	&  10 ($\times$10) & 0.9618 $\pm$ 0.0423 & 0.1189 $\pm$ 0.1021 & -394102.3 $\pm$ 11968.6 & 24 \\ \cline{2-6}
   	&  20 ($\times$10) & 0.9578 $\pm$ 0.0443 & 0.1344 $\pm$ 0.1220 & -394252.5 $\pm$ 11701.0 & 25 \\ \cline{2-6}
   	&  100 ($\times$10) & 0.9651 $\pm$ 0.0459 & 0.1065 $\pm$ 0.1136 & -394137.2 $\pm$ 12224.4 & 31 \\ \hline 
   	
    \end{tabular}
\end{center}
\vskip -1mm
\caption{Performance comparison of our two-round EM algorithm and the original EM algorithm for clustering cats, wolves and deers. }\label{table:CatDeerWolf} 
\vspace{-4mm}
\end{table*}

\textit{Evaluation metrics.} To evaluate the clustering quality, we use two metrics from \cite{Tuytelaars}: conditional purity and conditional entropy. Given the underlying ground-truth category labels $X$ (which are unknown to the algorithm) and the obtained cluster labels $Y$, the conditional purity is defined as the mean of the maximum category probabilities for $(X, Y)$, 
\[
\text{Purity}(X|Y) = \sum_{y\in Y} p(y) \max_{x\in X}p(x|y)
\]
and the conditional entropy is defined as, 
\[
\mathcal{H}(X|Y) = -\sum_{y\in Y} p(y) \sum_{x\in X}p(x|y)\log p(x|y)
\]
where both $p(y)$ and $p(x|y)$ are estimated on the training set, and we would expect higher purity and lower entropy for a better clustering algorithm.  

We then use the two-round EM algorithm to cluster the images and learn a binary template for each cluster. We compare with the original EM algorithm running for different numbers of iterations ($2, 10, 20, 100$ in the experiments) and starting with desired number $k$ of clusters (while the two-round EM starts with $l>k$ clusters and prunes them). A more robust EM could be obtained by starting with many random initialization and choosing the clustering result that has the largest log-likelihood. Such robust versions with different number of initializations are also evaluated in Tables \ref{table:CarBicycleMotorcycle}, \ref{table:CatDeerWolf} and \ref{table:all}. A ten-round version of the two-round EM (with eight additional EM iterations after the pruning step) is also evaluated.
The methods are evaluated in terms of conditional purity and conditional entropy. From the experiments one could see that the two-round EM algorithm can only be outperformed with a  five or ten random initializations of the standard EM algorithm. All the results are obtained based on 100 runs.

Fig. \ref{fig:CarBicycleMotorcycle} and \ref{fig:CatDeerWolf} show the results of two experiments (vehicles and animal faces). Table~\ref{table:CarBicycleMotorcycle} and \ref{table:CatDeerWolf} show the performance comparisons. Table \ref{table:all} shows the performance all data combined. In the learned templates, the existence of a sketch at each cell is represented by a bar at the center of this cell and at the orientation of the sketch. In each experiment, there are 15 images in each cluster, and the two-round EM is able to separate the clusters perfectly. For the real images, the templates are denser than those in Fig. \ref{fig:dog_examples} because the numbers of cells are larger. 

%%=============================================================================
%\begin{table*}
%\begin{center}
%    \begin{tabular}{ | l | l | l | l | l | }
%    \hline    
%   	\multicolumn{2}{|l|}{} & Cond. Purity & Cond. Entropy & Num$_{\text{perfect}}$ /100 \\ \hline
%   	\multicolumn{2}{|l|}{Two-round EM} &  \textbf{0.851}$\pm$ 0.095 & \textbf{0.323}$\pm$ 0.190 & \textbf{5} \\ \hline
%   	\multirow{4}{*}{Original EM} &  maxIter=2 & 0.751$\pm$0.121 & 0.509$\pm$ 0.223 & 0 \\ \cline{2-5}
%   	&  maxIter=10 & 0.803$\pm$0.129 & 0.360$\pm$0.229 & 3 \\ \cline{2-5}
%   	&  maxIter=20 & 0.803$\pm$0.129 & 0.360$\pm$0.229 & 3 \\ \cline{2-5}
%   	&  maxIter=100 & 0.803$\pm$0.129 & 0.360$\pm$0.229 & 3\\ \hline
%    \end{tabular}
%\end{center}
%\caption{Performance comparison between our two-round EM algorithm and the original EM algorithm for clustering cats, wolves, deers, motorcycles, bicycles and cars. In the table, Num$_{\text{perfect}}$ means the number of runs (out of the total 100 runs) which recover the underlying clusters perfectly. }\label{table:all} 
%\end{table*}

%=============================================================================
\begin{table*}[ht]
\begin{center}
    \begin{tabular}{ | l | l | l | l | l | l | }
    \hline    
   	Method & \parbox{1.5cm}{\#Round ($\times$\#Init)} & Cond. Purity & Cond. Entropy & Log-Likelihood & $N$ \\ \hline
   	\multirow{2}{*}{\parbox{1.5cm}{Tow-round EM}} 
   	& 2 ($\times$1)  & 0.8823 $\pm$ 0.0982 & 0.2526 $\pm$ 0.1891 & -401125.5 $\pm$ 28775.3 & 0 \\ \cline{2-6}
    & 10 ($\times$1) & 0.9030 $\pm$ 0.0911 & 0.1841 $\pm$ 0.1506 & -398672.2 $\pm$ 28364.9 & 6 \\ \hline
   	\multirow{12}{*}{\parbox{1.5cm}{Original EM}} 
   	&  2 ($\times$1) & 0.7389 $\pm$ 0.0995 & 0.5152 $\pm$ 0.1775 & -416130.8 $\pm$ 32791.7 & 0 \\ \cline{2-6}
   	&  10 ($\times$1) & 0.7744 $\pm$ 0.1160 & 0.4042 $\pm$ 0.2001 & -412488.2 $\pm$ 32696.2 & 2 \\ \cline{2-6}
   	&  20 ($\times$1) & 0.7961 $\pm$ 0.1129 & 0.3669 $\pm$ 0.1862 & -409443.2 $\pm$ 33581.6 & 2 \\ \cline{2-6}
   	&  100 ($\times$1) & 0.7883 $\pm$ 0.1265 & 0.3862 $\pm$ 0.2196 & -410602.3 $\pm$ 34139.1 & 0 \\ \cline{2-6} \cline{2-6}

   	&  2 ($\times$5) & 0.8468 $\pm$ 0.0769 & 0.3212 $\pm$ 0.1364 & -402958.6 $\pm$ 28555.4 & 1 \\ \cline{2-6}
   	&  10 ($\times$5) & 0.9082 $\pm$ 0.0857 & 0.1793 $\pm$ 0.1351 & -396123.3 $\pm$ 30142.8 & 9 \\ \cline{2-6}
   	&  20 ($\times$5) & 0.9158 $\pm$ 0.0790 & 0.1686 $\pm$ 0.1263 & -395430.3 $\pm$ 27609.8 & 8 \\ \cline{2-6}
   	&  100 ($\times$5) & 0.9022 $\pm$ 0.0854 & 0.1843 $\pm$ 0.1290 & -396627.6 $\pm$ 29355.7 & 5 \\ \cline{2-6} \cline{2-6}

   	&  2 ($\times$10) & 0.8836 $\pm$ 0.0746 & 0.2669 $\pm$ 0.1376 & -398549.9 $\pm$ 28767.4 & 1 \\ \cline{2-6}
   	&  10 ($\times$10) & 0.9483 $\pm$ 0.0602 & 0.1163 $\pm$ 0.0937 & -392213.1 $\pm$ 27651.5 & 10 \\ \cline{2-6}
   	&  20 ($\times$10) & 0.9504 $\pm$ 0.0633 & 0.1072 $\pm$ 0.0985 & -392517.4 $\pm$ 29030.8 & 16 \\ \cline{2-6}
   	&  100 ($\times$10) & 0.9574 $\pm$ 0.0566 & 0.0992 $\pm$ 0.0882 & -391457.3 $\pm$ 27788.0 & 10 \\ \hline 
   	
    \end{tabular}
\end{center}
\vskip -1mm
\caption{Performance comparison of our two-round EM algorithm and the original EM algorithm for clustering cats, wolves, deers, motorcycles, bicycles and cars.}\label{table:all} 
\vspace{-6mm}
\end{table*}

Currently we use a very simple sketch detector by thresholding the Gabor filter responses at different orientations. We will design more sophisticated features and associated detectors in future work.

\section{Discussion} 

This paper obtains theoretical guarantees on the performance of a two-round EM algorithm for learning mixture of Bernoulli templates, by generalizing the theory of \cite{dasgupta2000two}. Unlike the theoretical results for supervised learning, results on unsupervised learning such as clustering are relatively scarce. The results obtained in this paper can be useful for understanding the behavior of EM-type algorithms for unsupervised learning. 

In our future work, we shall improve the theoretical results by relaxing the conditions on the separation between the templates as well as the sample size. We shall also generalize Bernoulli templates to more general statistical models for images, such as templates with dependent switching of the binary components, as well as other non-Gaussian models such as exponential family models. 

\section{Acknowledgments}

The authors wish to acknowledge support from DARPA MSEE grant FA 8650-11-1-7149 and NSF grant DMS 1007889.
The authors thank Maria Pavlovskaia for her help with the experiment on synthetic sketches. Also thanks to Jianwen Xie for assistance with the animal face experiments, and to Prof. Song-Chun Zhu for valuable suggestions.

\section*{Appendix: Proofs}

{\em Proof of Prop. \ref{thm:expBern}.}

1. We have
\[
\begin{split}
&E[D(\bx,\bP)]=E[\sum_{k=0}^n B_k]=\sum_{k=0}^n E[B_k]=nq
\end{split}
\]
and
\[
\begin{split}
E[D(&\bx,\bP)^{2}]=E[(\sum_{k=0}^n B_k)^{2}]=E[\sum_{i=0}^n B_i^{2}+\sum_{i\not =j} B_i B_j]\\
&=\sum_{i=0}^n E[B_i]+\sum_{i\not =j} E[B_i B_j]=nq+n(n-1)q^{2}\\
Var&(D(\bx,\bP))=E[D(\bx,\bP)^{2}]-E[D(\bx,\bP)]^{2}\\
&=n(n-1)q^{2}+nq-n^{2}q^{2}=nq(1-q)
\end{split}
\]
2. Let $d=D(\bP,\by)$. Without loss of generality, let $\bP=(\bA,\bB), \by=(\bA,1-\bB)$ where $\bB\in \{0,1\}^d$ and $\bx=(\bu,\bz),\bu\sim \bA,\bz\sim \bB$. 
Observe that if two random variables are independent then $Var(A+B)=Var(A)+Var(B)$.
Then
\[
\begin{split}
E[D(\bx,\by)]&=E[D(\bu,\bA)+D(\bz,1-\bB)]\\&=(n-d)q+(d-E[D(\bz,\bB)])=(n-d)q+d-dq\\
Var(D(\bx,\by))&=Var[D(\bu,\bA)+d-D(\bz,\bB)]\\
&= Var[D(\bu,\bA)]+Var[d-D(\bz,\bB)]\\
&=(n-d)q(1-q)+dq(1-q)=nq(1-q)
\end{split}
\]
3. In the case when $\bx,\by \sim \bP$ we have 
\[
\begin{split}
E_{\bx,\by}[&D(\bx,\by)]=E_\bx[E_\by[D(\bx,\by)]]=E_\bx[nq+D(\bx,\bP)(1-2q)]\\
&=nq+nq(1-2q)=2nq(1-q)\\
Var_{\bx,\by}&(D(\bx,\by))=E_{\bx,\by}[D(\bx,\by)^{2}]-\hspace{-0.5mm}(E_{\bx,\by}[D(\bx,\by)])^{2}\\
&=E_\bx(E_\by[D(\bx,\by)^{2}])-E_\bx(E_\by^{2}[D(\bx,\by)])\\&+E_\bx(E_\by^{2}[D(\bx,\by)])-(E_\bx[E_\by(D(\bx,\by))])^{2}\\
&=E_\bx(Var_\by[D(\bx,\by)])+Var_\bx[E_\by(D(\bx,\by))]\\
&=E_\bx(nq(1-q))+Var_\bx[nq+D(\bx,\bP)(1-2q)]\\
&=nq(1-q)+nq(1-q)(1-2q)^{2}
\end{split}
\]

4.  In the case when $\bx\sim \bP, \by \sim \bQ$ we have 
\[
\begin{split}
E_{\bx,\by}[D&(\bx,\by)]=E_\bx[E_\by[D(\bx,\by)]]
=E_\bx[nq+D(\bx,\bQ)(1-2q)]\\
&=nq+(nq+D(\bP,\bQ)(1-2q))(1-2q)\\
&=2nq(1-q)+D(\bP,\bQ)(1-2q)^{2}\\
Var_{\bx,\by}&(D(\bx,\by))=E_\bx(Var_\by[D(\bx,\by)])+Var_\bx[E_\by(D(\bx,\by))]\\
&=E_\bx(nq(1-q))+Var_\bx[nq+D(\bx,\bQ)(1-2q)]\\&=nq(1-q)+nq(1-q)(1-2q)^{2}. \Box
\end{split}
\]

{\em Proof of Prop. \ref{thm:devBern}.}
Statements a), b) follow directly from the Chernoff inequality. 

c) Let $C$ be indices of the $n-d$ common elements of $\bP$ and $\bQ$. 
Let $B_i$ be the Bernoulli event that the $i$-th element of $\bx$ and $\by$ are different. Then $E(B_i)=2q(1-q)$ if $i\in C$ and  $E(B_i)=q^{2}+(1-q)^{2}$ if $i\not \in C$. Observe that $D(\bx,\by)=\sum_{i=1}^n B_i$. Thus by the Chernoff inequality, since $\nu=E[D(\bx,\by)]=2nq(1-q)+d(1-2q)^{2}$ we get
\[
\bP(|D(\bx,\by)-\nu|>\epsilon \nu )\leq 2e^{-\nu \epsilon^{2}/3}. \; \Box
\]

{\em Proof of Prop. \ref{prop:eps}.}
a) From point c) of Prop. \ref{thm:devBern} with $\bP=\bQ$, we have $\nu=\nu(\bP,\bP)=2nq(1-q)$ so for any two points $\bx,\by\in S_i$ we have $
\bP(|D(\bx,\by)-\nu|>\nu\epsilon_0 /\sqrt{q})\leq 2e^{-\nu \epsilon_0^{2}/3q}$. Thus for all $m(m-1)/2$ combinations of two points we have 
\[
\begin{split}
\bP(|D(\bx,\by)-\nu|>\nu\epsilon_0/\sqrt{q} )&\leq m(m-1)e^{-\nu \epsilon_0^{2}/3q}
<m^{2}e^{-2n(1-q) \epsilon_0^{2}/3}
\end{split}
\]
b) Similar to the proof of a), with $\nu=\nu(\bP,\bQ)=2nq(1-q)+d(\bP,\bQ)(1-2q)^{2}=2nq(1-q)+nc_{ij}(1-2q)^{2} \geq n\min(c,0.5)$. We obtain
\[
\bP(|D(\bx,\by)-\nu|>\nu\epsilon_0 )<m^{2}e^{-n\min(c,0.5) \epsilon_0^{2}/3}
\]
c) From point b) of Prop. \ref{thm:devBern} we have $\bP(|D(\bx,\bP_i)-nq|>\epsilon_0 n\sqrt{q})\leq 2e^{-n \epsilon_0^{2}/3}$ so for all $m$ points we have
\[\bP(|D(\bx,\bP_i)-nq|>\epsilon_0 n\sqrt{q})\leq 2me^{-n \epsilon_0^{2}/3}\]

Similarly,  we have 
\[
\begin{split}
\bP(|D(\bx,\bP_j)-&n(q+c_{ij}(1-2q))|>\epsilon_0 n(q+c_{ij}(1-2q)))\\
&\leq 2e^{-n(q+c_{ij}(1-2q)) \epsilon_0^{2}/3}\leq 2e^{-n \min(c,0.5) \epsilon_0^{2}/3}
\end{split}\] so for all $m$ points we have
\[
\begin{split}
\bP(|D(\bx,\bP_j)-&n(q+c_{ij}(1-2q))|>\epsilon_0 n(q+c_{ij}(1-2q)))\\
&\leq 2me^{-n \min(c,0.5) \epsilon_0^{2}/3}
\end{split}\]

d)  Let $B_j$ be Bernoulli event that sample $j$ is drawn from template $\bP_i$. Then $E[B_j]=w_i$ and  from the Chernoff bound
\[
\begin{split}
\bP(|S_i|< \frac{1}{2}mw_i)&=\bP\left (\frac{\sum_{j=1}^m B_j}{m}<w_i(1-  \frac{1}{2})\right )\\
&< e^{-m w_i (1/2)^{2}/2}< e^{-m w_{min}/8}.\Box
\end{split}
\]

{\em Proof of Lemma \ref{lem:absZq}.} 
The mean of $mn$ Bernoullis $B_{ij}$ with $E[B_{ij}]=q$ (the coordinates of the $Z_i$) satisfies
\[
\bP(\frac{\sum B_{ij}}{mn}-q>\epsilon q)<e^{-mnq \epsilon^{2}/3}
\]
So %each thresholded component $\bT_i$ of the mean $\bT$ is a Bernoulli with $E[\bT_i]\leq A$. Then from Chernoff inequality we have
\[
\bP(\sum_{i=1}^n Z_i -nq >\epsilon nq)\leq e^{-mnq\epsilon^{2}/3}
\]
and we take $\epsilon=\lambda/nq. \Box$

{\em Proof of Prop. \ref{prop:avg1}.} First, it is sufficient to prove it for subsets of size exactly $t$, otherwise we increase $t$. Without loss of generality, we can assume $\bP=\mathbf{0}$. From Lemma \ref{lem:absZq} we have 
\[
\bP(D(\mu,\bP)-nq>\lambda)\leq e^{-t\lambda^{2}/3nq}
\]
The number of $t$-point subsets of $S_1$ is ${m\choose t} <(me/t)^t$, thus
\[
\begin{split}
\bP(\exists \text { subset of $t$ points s.t. }  &D(\mu,\bP)-nq>\lambda)
\leq \left ( \frac{me}{t}\right )^ke^{-t\lambda^{2}/3nq}%\leq \left ( \frac{me}{t}\right )^ke^{-nkq\lambda^{2}/6}
\end{split}
\]
%where the last inequality holds when $\lambda \geq 4$.
Solving for $ \left ( \frac{me}{t}\right )^ke^{-t\lambda^{2}/3nq}=\delta$ we get
\[
\lambda = \sqrt{\frac{3nq}{t} \left (t\ln \frac{me}{t}+\ln \frac{1}{\delta} \right )}
\]
therefore \[
\begin{split}
\bP\bigg{[}\exists \text { subset of $t$ points s.t. }  D(\mu,\bP)-nq
>\sqrt{\frac{3nq}{t} \left (t\ln \frac{me}{t}+\ln \frac{1}{\delta} \right )}\bigg{]}
\leq \delta. \;\Box
\end{split}
\]

{\em Proof of Prop \ref{prop:wtavg}.}
Sort the points $\bx\in S$ by $D(\bx,\bP)=\sum_{i=1}^n |\bx_i-\bP_i|$ and take $T$ as the ones with $ |T|=\lfloor \sum_{\bx\in S} w_\bx\rfloor$ largest values.
Then
\[
\sum_{\bx\in T} \sum_{i=1}^n |\bx_i-\bP_i| \geq \sum_{\bx\in S} w_\bx \sum_{i=1}^n |\bx_i-\bP_i|
\]
so 
\[
\begin{split}
D(\mu_T,\bP)&=\hspace{-1mm}\sum_{i=1}^n \frac{\sum_{\bx\in T} |\bx_i-\bP_i|}{|T|} \\&\geq \sum_{i=1}^n \frac{\sum_{\bx\in S} w_\bx |\bx_i-\bP_i|}{|T|}\\
&\geq \sum_{i=1}^n \frac{\sum_{\bx\in S} w_\bx |\bx_i-\bP_i|}{\sum_{\bx\in S} w_\bx}=D(\mu_w,\bP).\; \Box
\end{split}
\]

{\em Proof of Prop. \ref {prop:em0}.} Let $B_i$ be the Bernoulli event that a random sample from the mixture comes from the $i$-th true template $\bP_i$. Then $E[B_i]=w_i$. Having $l$ random samples $B_{ij}$ from the Bernoulli event $B_i$, then 
\[
\begin{split}
\bP(\sum_{j=1}^l B_{ij}\leq 1)&=(1-w_i)^l+lw_i(1-w_i)^{l-1}\\
&\leq (1+l)(1-w_{min})^l\leq (l+1)e^{-lw_{min}}
\end{split}
\] 
so $\bP(\sum_{j=1}^l B_{ij}\geq 2)\geq 1-(l+1)e^{-lw_{min}}$.
Thus $\bP(\bP_i \text{ is represented twice})\geq 1-(l+1)e^{-lw_{min}}$, so $\bP(\bP_i \text{ is represented twice},  \forall i=\overline{1,k})\geq (1-(l+1)e^{-lw_{min}})^k\geq 1-k(l+1)e^{-lw_{min}}$.

2. From Chernoff bound we have $\bP(\sum_{j=1}^l B_{ij}>15/8 lw_i)<e^{-lw_i (7/8)^{2}/3}<e^{-lw_i /4}$, which implies the results.

3. As there exist $\bT_i',\bT_j'$ representing the same cluster, then $2nq_0(1-q_0)\leq D(\bT_i',\bT_j')\leq 2n(1-q)(q+ \epsilon_0\sqrt{q})$ (from Prop. \ref{prop:eps}, a). 
Also from Prop. \ref{prop:eps}, if the minimum is attained for two centers  $\bT_i',\bT_j'$ representing the same cluster, we are done. Otherwise
\[
\begin{split}
2nq_0(1-q_0)&=(2nq(1-q)+nc_{ij}(1-2q)^{2})(1\pm \epsilon_0)\\
&\geq 2nq(1-q)(1-\epsilon_0)\geq 2n(1-q)(q-\epsilon_0 \sqrt{q})
\end{split}
\]
so both parts of the inequality are proved. $\Box$
 
{\em Proof of Prop \ref{prop:probs} .}
We have
\[
\frac{p_{i'}^{(1)}(\bx)}{p_{j'}^{(1)}(\bx)}=\frac{q_0^{D(\bx,\bT_{i'}^{(0)})}(1-q_0)^{n-D(\bx,\bT_{i'}^{(0)})}}{q_0^{D(\bx,\bT_{j'}^{(0)})}(1-q_0)^{n-D(\bx,\bT_{j'}^{(0)})}}=a^{D(\bx,\bT_{j'}^{(0)})-D(\bx,\bT_{i'}^{(0)})},
\]
with $a=\frac{1-q_0}{q_0}>1$. 
But from Prop. \ref{prop:eps}
\[
\begin{split}
D(\bx,\bT_{j'}^{(0)})-&D(\bx,\bT_{i'}^{(0)})>(2nq(1-q)+nc_{ij}(1-2q)^{2})(1-\epsilon_0)\\&-2n(1-q)(q+\epsilon_0\sqrt{q})\\
&=-2n(q+\sqrt{q})(1-q)\epsilon_0
+nc_{ij}(1-2q)^{2}(1-\epsilon_0)\\&>nc_{ij}(1-2q)^{2}/2
\end{split}
\]
since we have the following condition 
\[
c(1-2q)^2(\frac{1}{2}-\epsilon_0)\geq 2\epsilon_0(1-q)(q+\sqrt{q})
\]
obtained from $\epsilon_0\leq E$. 
We also have since $\epsilon_0<1/4$
\[
\begin{split}
a&=\frac{(1-q_0)^{2}}{q_0(1-q_0)}\geq \frac{1/4}{(1-q)(q+\epsilon_0\sqrt{q})} \\&\geq \frac{1}{4 (1-q) (q+1/4\sqrt{q})}=\frac{1}{4(1-q) (4q+\sqrt{q})}
\end{split}
\]
So
\[
\begin{split}
\frac{p_{i'}^{(1)}(\bx)}{p_{j'}^{(1)}(\bx)}&\geq \exp (\frac{n}{2}c_{ij}(1-2q)^{2}\ln \frac{1}{4(1-q) (4q+\sqrt{q})})\\&=\exp(nc_{ij}B(1-2q)).\Box
\end{split}
\]

{\em Proof of Prop. \ref{prop:em1}.} Without loss of generality we can assume $\bP_i=0$.
\[
\begin{split}
D(\bT_{i'}^{(1)}&,\bP_i)=\frac{\sum_{k=1}^n \sum_\bx p_{i'}^{(1)}(\bx) \bx_k}{\sum_\bx p_{i'}^{(1)}(\bx)}\\
&\leq \frac{\sum_{k=1}^n\sum_{\bx\in S_i} p_{i'}^{(1)}(\bx) \bx_k}{\sum_\bx p_{i'}^{(1)}(\bx)}+ \frac{\sum_{k=1}^n\sum_{\bx\not \in S_i} p_{i'}^{(1)}(\bx)\bx_k}{\sum_\bx p_{i'}^{(1)}(\bx)}\\
&\leq \frac{\sum_{k=1}^n\sum_{\bx\in S_i} p_{i'}^{(1)}(\bx) \bx_k}{\sum_{\bx\in S_i} p_{i'}^{(1)}(\bx)}+\frac{\sum_{j\not =i}\sum_{\bx \in S_j} p_{i'}^{(1)}(\bx) D(\bx,0)}{\sum_\bx p_{i'}^{(1)}(\bx)}
\end{split}
\]
From Prop \ref{prop:probs}, for any $\bx\in S_j, j\not =i$  we have $p_{i'}^{(1)}(\bx)\leq e^{-nc_{ij} B(1-2q)}\leq  e^{-nc B(1-2q)}$. Then
\[
\begin{split}
\sum_{\bx\in S_i} p_{i'}^{(1)}(\bx)&\geq \sum_\bx  p_{i'}^{(1)}(\bx) - \sum_{j\not =i}\sum_{\bx \in S_j} p_{i'}^{(1)}(\bx)\\
&\geq 
 mw_T-me^{-ncB(1-2q)}\geq mw_T/4+1
\end{split}
\]
from  $w_T= 1/4l $ and conditions $m\geq 8l$ (C2) and 
$ncB(1-2q)\geq \ln (16l)$ (C1).

From Prop. \ref{prop:wtavg} there exists $T\subset S_i$ with $|T|=\lfloor mw_T/4+1\rfloor$ such that $D(\mu_T,0)\geq D(\mu_w,0)$. From Prop \ref{prop:avg1}, with probability $1-1/n$
\begin{equation}
\begin{split}
&\frac{\sum_{j=1}^n\sum_{\bx\in S_i} p_{i'}^{(1)}(\bx) \bx_j}{\sum_{\bx\in S_i} p_{i'}^{(1)}(\bx)}\leq D(\mu_T,0)
\leq nq+\sqrt{3nq\left( \ln \frac{4|S_i|e}{mw_T}+\frac{4}{mw_T}\ln n \right )} 
\end{split}
\end{equation}
Then since $1/w_T=4l$ we have
\begin{equation}
\begin{split}
&\frac{\sum_{j=1}^n\sum_{\bx\in S_i} p_{i'}^{(1)}(\bx) \bx_j}{\sum_{\bx\in S_i} p_{i'}^{(1)}(\bx)}
 \leq  nq+ \sqrt{3nq(\ln 16el +\frac{16l}{m} \ln n)} 
\leq nq+\sqrt{6nql}  \label{eq:ppart1}
\end{split}
\end{equation}from condition $m>16\ln n$ (C2) and $\ln 16el<l$ (which holds for $l
\geq 9$).

For the second term, from Prop \ref{prop:eps} we have, for $\bx\in S_j$
\[
D(\bx,\bP_i)\leq (nq+nc_{ij}(1-2q))(1+\epsilon_0)
\]
where since $\epsilon_0\leq 0.5$ we have
\[
\begin{split}
p_{i'}D(\bx,\bP_i)&\leq e^{-nc_{ij}B(1-2q)}(nq+nc_{ij}(1-2q))(1+\epsilon_0)
\\&\leq  e^{-nc_{ij}B(1-2q)/2}\leq e^{-ncB(1-2q)/2}
\end{split}
\] so
\begin{equation}
\begin{split}
\frac{\sum_{j\not =i}\sum_{\bx \in S_j} p_{i'}^{(1)}(\bx) D(\bx,\bP_i)}{\sum_\bx p_{i'}^{(1)}(\bx)}\hspace{-1mm}&\leq \hspace{-1mm}\frac{1}{mw_T}\hspace{-1mm}\sum_{j\not =i}\hspace{-1mm}\sum_{\bx \in S_j} p_{i'}^{(1)}(\bx) D(\bx,\bP_i) 
\\&\leq \frac{1}{w_T}e^{-ncB(1-2q)/2}<\sqrt{6nql} \label{eq:ppart2}
\end{split}
\end{equation}
using condition $ncB(1-2q)\geq \ln (8l/3nq)$ (C1).
Putting together \eqref{eq:ppart1} and \eqref{eq:ppart2} we get the result.
$\Box$

{\em Proof of Prop. \ref{prop:pruning}. } a). From Proposition \ref{prop:eps} and \ref{prop:em0} we have that $|S_i|>mw_i/2$ and at most $15lw_i/8$ initial centers are from $S_i$.

Let $i'$ be such that $\bT_{i'}^{(0)} \in S_i$ and $\bx\in S_i$. For any $j$ such that $\bT_j^{(0)}\not \in S_i$ we have from Prop \ref{prop:probs} $p_{i'}^{(1)}(\bx)/p_j^{(1)}(\bx)\geq e^{nc_{ij}B(1-2q)}\geq e^{ncB(1-2q)}$. Then $p_j^{(1)}(\bx)\leq e^{-ncB(1-2q)}$ and thus $\sum_{k,\bT_k^{(1)}\in S_i}p_k^{(1)}(\bx)\geq 1-le^{-ncB(1-2q)}$. But then
\[
\begin{split}
\sum_{k,\bT_k^{(1)}\in S_i} w_k^{(1)}&=\frac{\sum_{\bx\in S}\sum_{k,\bT_k^{(1)}\in S_i} p_k^{(1)}(\bx)}{m}
\\&\geq\frac{|S_i|(1-le^{-ncB(1-2q)})}{m}\geq \frac{ w_i}{2} (1-le^{-ncB(1-2q)})
\end{split}
\]
But $|\{j,\bT_j^{(1)}\in S_i\}|\leq 15lw_i/8$ so there is a $j, \bT_j^{(1)}\in S_i$ such that
\[
\begin{split}
w_j^{(1)}&\geq  \frac{w_i(1-le^{-ncB})/2}{15lw_i/8}=\frac{1-le^{-ncB(1-2q)}}{15l/4}
\\&\geq \frac{1}{4l}=w_T
\end{split}
\]
using condition $ncB(1-2q)\geq \ln (16l)$ (C1), thus $C_i$ is not empty.

b) Pick any $\bT_{i'}^{(1)}\in C_i$ and $\bT_{j'}^{(1)},\bT_{j"}^{(1)}\in C_j$ for $i\not =j$. Then from Proposition \ref{prop:em1} we have
\[
D(\bT_{j'}^{(1)},\bT_{j"}^{(1)})\leq 2nq+4\sqrt{6nql}
\]
while using Proposition \ref{prop:em1} and the triangle inequality we get 
\[
\begin{split}
&D(\bT_{i'}^{(1)},\bT_{j'}^{(1)})\geq D(\bP_i,\bP_j)- 2nq-4\sqrt{6nql} \geq \\
&\geq nc -2nq-4\sqrt{6nql}>2nq+4\sqrt{6nql}
\end{split}
\]
from condition $c\geq 4q+8\sqrt{6ql/n}$ (C3), so we can take $\tau=\frac{1}{2}n c$.

c)  There are $k$ true clusters, exactly as many as selected templates. If two selected templates were from the same cluster, there should be a cluster that has no selected templates. But the two templates from the same cluster are at distance at most $\tau$ while the distance of a template from the unselected cluster has distance more than $\tau$, we get a contradiction.
$\Box$

{\em Proof of Prop. \ref{prop:probs2} .}
Using the triangle inequality, Prop. \ref{prop:eps} and Prop. \ref{prop:em1} we have
\[
\begin{split}
D(\bx,\bT_i^{(1)})&\leq D(\bx,\bP_i)+ D(\bT_i^{(1)},\bP_i)
\leq n(q+\epsilon_0\sqrt{q})+nq+2\sqrt{6nql}
\end{split}
\]
and
\[
\begin{split}
&D(\bx,\bT_j^{(1)})\geq  D(\bx,\bP_j)-D(\bT_j^{(1)},\bP_j)
\\&\geq n(q+c_{ij}(1-2q))(1-\epsilon_0)-nq-2\sqrt{6nql},
\end{split}
\]
so
\[
\frac{p_{i}^{(2)}(\bx)}{p_{j}^{(2)}(\bx)}=\frac{q_0^{D(\bx,\bT_{i}^{(1)})}(1-q_0)^{n-D(\bx,\bT_{i}^{(1)})}}{q_0^{D(\bx,\bT_{j}^{(1)})}(1-q_0)^{n-D(\bx,\bT_{j}^{(1)})}}=a^{D(\bx,\bT_{j}^{(1)})-D(\bx,\bT_i^{(1)})},
\]
where $a=\frac{1-q_0}{q_0}>1$, and therefore
\[
\begin{split}
\frac{p_{i}^{(2)}(\bx)}{p_{j}^{(2)}(\bx)}&\geq \exp([n(q+c_{ij}(1-2q))(1-\epsilon_0)-n(q+\epsilon_0\sqrt{q})-
-2nq-4\sqrt{6nql}]\ln a)\\
&=\exp(n[c_{ij}(1-2q)(1-\epsilon_0)-2q-\epsilon_0(q+\sqrt{q}) 
-4\sqrt{\frac{6ql}{n}}  ]\ln a) \\&\geq  \exp(nc_{ij}\frac{1}{4}(1-2q)\ln \frac{1}{6\sqrt{q}})
\end{split}
\]
using the condition
\[
c(1-2q)(\frac{3}{4}-\epsilon_0)\geq 2q+\epsilon_0(q+\sqrt{q})+4\sqrt{\frac{6ql}{n}}
\]
obtained from $\epsilon_0\leq t$. $\Box$

{\em Proof of Theorem \ref{thm:em2}.}
First we compute the probability that the theorem holds.

Proposition \ref{prop:eps} holds with probability at least $1-m^{2}e^{-2n(1-q) \epsilon_0^{2}/3}$$-m^{2}e^{-n\min(c,0.5) \epsilon_0^{2}/3}-2me^{-n\epsilon_0^{2}/3}-2me^{-n\min(c,0.5)\epsilon_0^{2}/3}-ke^{-mw_{min}/8}$.
Proposition \ref{prop:em0} holds with probability at least $1-k(l+1)e^{-lw_{min}}-ke^{lw_{min}/4}$.
Proposition \ref{prop:em1} holds with probability at least $1-1/n$ for each of the $k$ clusters. All other propositions hold if these three propositions hold. 

Thus with probability  $1-m^{2}e^{-2n(1-q) \epsilon_0^{2}/3}
-m^{2}e^{-n\min(c,0.5) \epsilon_0^{2}/3}-2me^{-n\epsilon_0^{2}/3}-2me^{-n\min(c,0.5)\epsilon_0^{2}/3}-ke^{-mw_{min}/8}-k(l+1)e^{-lw_{min}}-ke^{lw_{min}/4}-k/n$ all propositions hold for all clusters.

Now we prove the distance inequality. Similar to the proof of Proposition \ref{prop:probs} we have
\[
\begin{split}
D&(\bT_{i}^{(2)},\bP_i)=D(\frac{\sum_\bx p_{i}^{(2)}(\bx) \bx}{\sum_\bx p_{i}^{(2)}(\bx)},\bP_i)=\frac{\sum_\bx p_{i}^{(2)}(\bx) D(\bx,\bP_i)}{\sum_\bx p_{i}^{(2)}(\bx)}
\\&\leq \frac{\sum_{\bx\in S_i} p_{i}^{(2)}(\bx) D(\bx,\bP_i)}{\sum_{\bx\in S_i} p_{i}^{(2)}(\bx)}+\frac{\sum_{j\not =i}\sum_{\bx \in S_j} p_{i}^{(2)}(\bx) D(\bx,\bP_i)}{\sum_{\bx\in S_i} p_{i}^{(2)}(\bx)}
\end{split}
\]

From Proposition \ref{prop:probs2} we have for $\bx\in S_i$, $p_j^{(2)}(\bx)\leq p_i^{(2)}(\bx) e^{-nc B/2}\leq e^{-nc B/2}$ so 
\[
p_i^{(2)}(\bx)= 1-\sum_{j\not =i}p_j^{(2)}(\bx)\geq 1-ke^{-nc B/2}
\]

So the first term is bounded as:
\[
\begin{split}
&\frac{\sum_{\bx\in S_i} p_{i}^{(2)}(\bx) D(\bx,\bP_i)}{\sum_{\bx\in S_i} p_{i}^{(2)}(\bx)}\leq \frac{\sum_{\bx\in S_i} (1-ke^{-nc B/2}) D(\bx,\bP_i)}{|S_i|(1-ke^{-nc B/2})}
\\&+\frac{\sum_{\bx\in S_i} (p_{i}^{(2)}(\bx)-(1-ke^{-nc B/2}) )D(\bx,\bP_i)}{\sum_{\bx\in S_i} p_{i}^{(2)}(\bx)}\\
&\leq D(\text{mean}(S_i),\bP_i)+\frac{\sum_{\bx\in S_i} ke^{-nc B/2}D(\bx,\bP_i)}{|S_i|(1-ke^{-nc B/2})}
\\&\leq D(\text{mean}(S_i),\bP_i)
+\frac{|S_i| ke^{-nc B/2}n(q+\epsilon\sqrt{q})}{|S_i|(1-ke^{-nc B/2})}\\
&\leq  D(\text{mean}(S_i),\bP_i)+2ke^{-nc B/2}nq
\end{split}
\]
when $\epsilon<\sqrt{q}(1-2ke^{-nc B/2})$.

The second term is bounded as:
\[
\begin{split}
&\frac{\sum_{j\not =i}\sum_{\bx \in S_j} p_{i}^{(2)}(\bx) D(\bx,\bP_i)}{\sum_{\bx\in S_i} p_{i}^{(2)}(\bx)} \\&\leq \frac{mn e^{-nc B/2}}{|S_i|(1-ke^{-nc B/2})}
\leq \frac{2nq e^{-nc B/4}}{w_i(1-ke^{-nc B/2})}\leq \frac{3}{w_i}nq e^{-nc B/4}
\end{split}
\]
when $e^{-nc B/8}<q$ and $ke^{-nc B/2}<1/3$.

From the inequality 
\[
ke^{-nc B/4}\leq 1\leq \frac{1}{w_{min}}
\] we get the result. $\Box$

\end{document}